\documentclass{article}

\usepackage{microtype}
\usepackage{graphicx}
\usepackage{subfigure}
\usepackage{booktabs} 

\usepackage{hyperref}

\usepackage[accepted]{icml2023}

\usepackage{amsmath}
\usepackage{amssymb}
\usepackage{mathtools}
\usepackage{amsthm}

\usepackage[capitalize,noabbrev]{cleveref}

\theoremstyle{plain}

\theoremstyle{definition}

\theoremstyle{remark}

\usepackage[textsize=tiny]{todonotes}

\icmltitlerunning{PLay: Parametrically Conditioned Layout Generation}

\begin{document}

\twocolumn[
\icmltitle{PLay: Parametrically Conditioned Layout Generation using Latent Diffusion}




\begin{icmlauthorlist}
\icmlauthor{Chin-Yi Cheng}{comp}
\icmlauthor{Forrest Huang}{comp}
\icmlauthor{Gang Li}{comp}
\icmlauthor{Yang Li}{comp}
\end{icmlauthorlist}

\icmlaffiliation{comp}{Google Research, Mountain View, United States}

\icmlcorrespondingauthor{Chin-Yi Cheng}{cchinyi@google.com}
\icmlcorrespondingauthor{Yang Li}{liyang@google.com}

\icmlkeywords{Machine Learning, ICML}

\vskip 0.3in
]



\printAffiliationsAndNotice{}  

\begin{abstract}
Layout design is an important task in various design fields, including user interface, document, and graphic design. As this task requires tedious manual effort by designers, prior works have attempted to automate this process using generative models, but commonly fell short of providing intuitive user controls and achieving design objectives. In this paper, we build a conditional latent diffusion model, PLay, that generates parametrically conditioned layouts in vector graphic space from user-specified guidelines, which are commonly used by designers for representing their design intents in current practices. Our method outperforms prior works across three datasets on metrics including FID and FD-VG, and in user study. Moreover, it brings a novel and interactive experience to professional layout design processes.

\end{abstract}

\section{Introduction}
\label{intro}
Layouts are important artifacts that represent the design and arrangements of their encapsulated elements. They are used extensively in creative fields ranging from design to engineering, supporting the authoring processes of numerous downstream products, such as user interfaces (UIs), documents, posters, architectural floorplans, and even printed circuit boards (PCBs). Designing a good layout requires thorough consideration of different aspects, such as the function, aesthetics, and domain-specific conditions and rules. Moreover, many of these aspects and objectives cannot be easily and explicitly evaluated and computed. Therefore, layout design has been a time-consuming, iterative, and manual process.

\begin{figure}[ht]
\begin{center}
\centerline{\includegraphics[width=0.64\columnwidth]{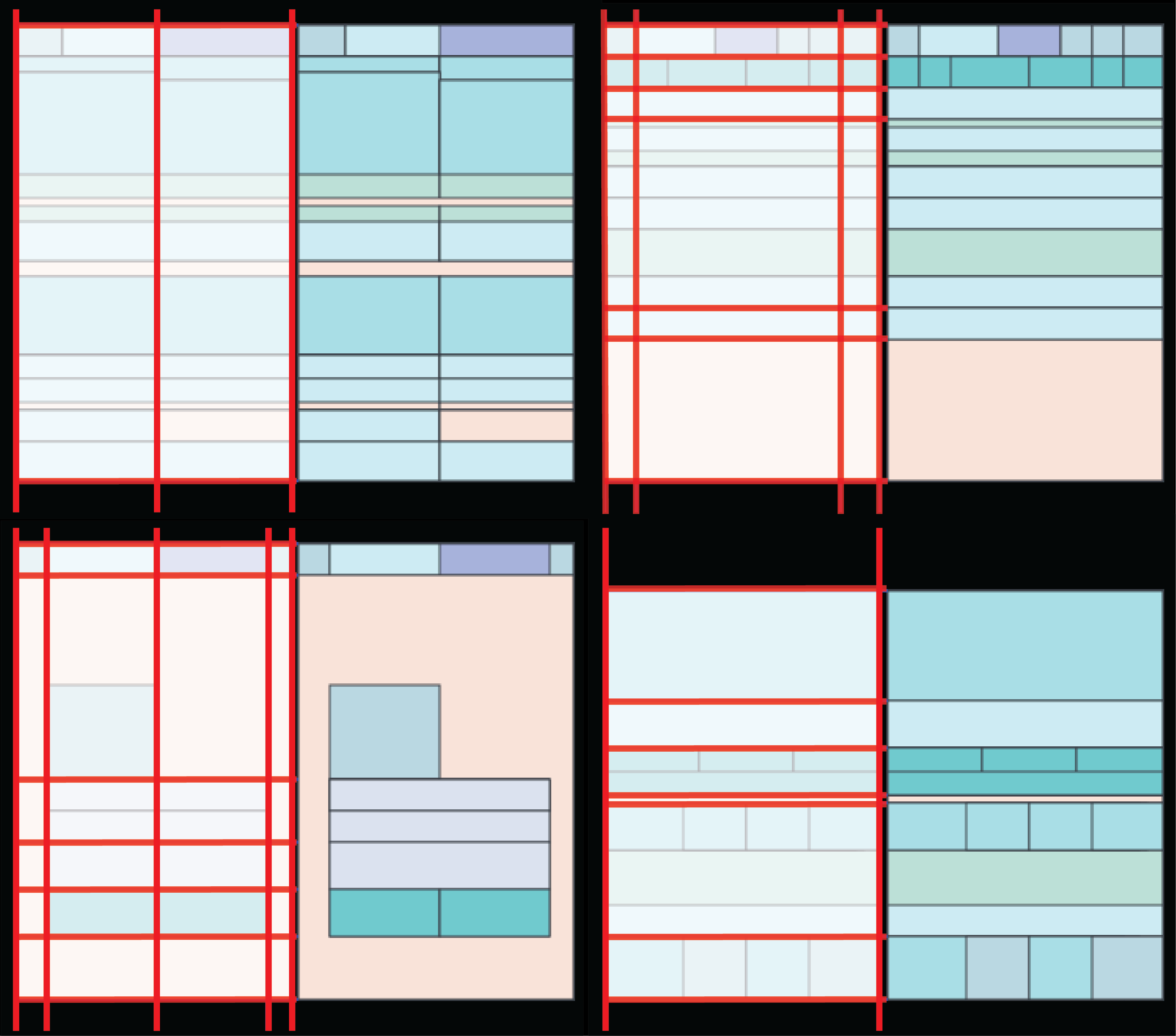}}
\caption{Sample results. The red lines on the left of each example represent the guideline conditions, and the generated layout is on the right. The generated layouts are also placed under the guidelines to visualize their alignments to the guidelines.}
\label{teaser}
\end{center}
\vskip -0.2in
\end{figure}

Several recent works have tried to improve the layout design process by automating it using generative models. However, these models are typically unconditional, making them difficult for users to adopt in realistic applications. A large number of unconditionally, randomly generated layouts are not useful in practice since it requires users to manually check each of them on whether the original objectives and constraints have been met. More importantly, for users to enjoy co-designing with generative models and trust the generated layouts, they need to be able to express their design ideas and drive the generation process \cite{musicco}. Therefore, a model conditioned on user inputs would help their adoption in users' workflows, and the type of supported inputs should represent design intention and constraints. 

The key towards achieving our goal of conditional generation that can be widely adopted is to choose an adequate type of condition for generating design layouts. Different types of conditions have been explored in image-based generative models, such as text \cite{imagen, ldm, dalle2} and semantic segments \cite{gaugan}, but these might not be ideal choices for layout generation. Text conditions have shown to be powerful for artistic purposes, but they cannot provide exact and detailed control critical for design. On the other hand, conditioning on semantic segments allows for precise control, but is tedious and time-consuming for users to manually author all segments in the scene, which is equivalent to hand-drawing the entire layout in our case of layout design.

\begin{figure}[ht]
\begin{center}
\centerline{\includegraphics[width=0.95\columnwidth]{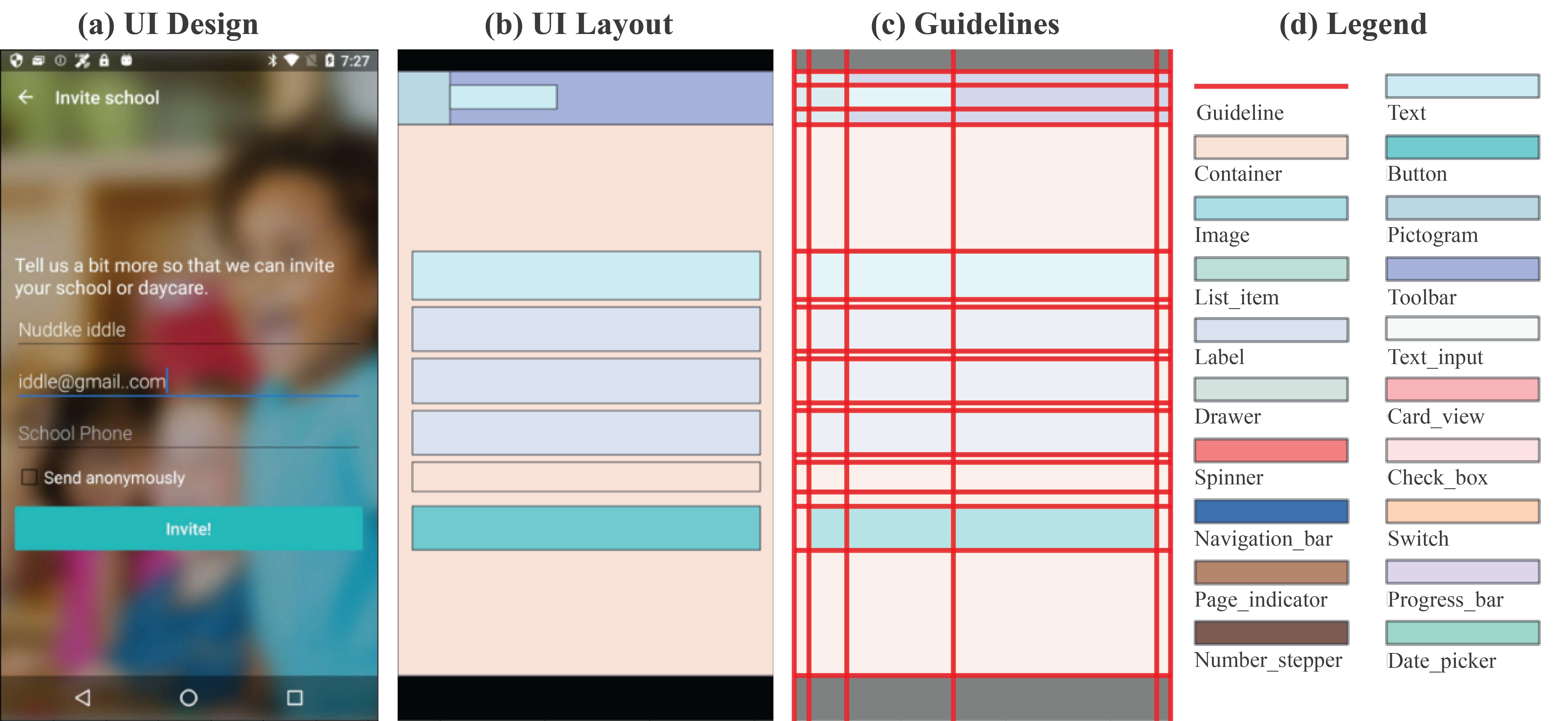}}
\caption{Layout and guidelines. A layout (b) of a UI design (a) consists of vector graphic elements with types and geometric properties. Guidelines (c) can represent the ideas and constraints.}
\label{ui_data_gl}
\end{center}
\vskip -0.2in
\end{figure}

In this work, we investigate layout design workflows in several domains, including UI, document, and architectural design, and identify a widely used representation: \textit{guidelines} (e.g., the red lines in Figure~\ref{ui_data_gl}c), as conditions. Guidelines, also referred to as guides, grids, or partitions, are a set of lines that serves multiple purposes, for instance: partitioning the design space for preferred proportions and alignments; expressing design ideas such as the two-column style (top-left example, Figure~\ref{teaser}); and showing rules such as paddings, margins, and gaps between elements. However, a common pain point for designers is that changing guidelines (which express design thoughts) would require manually redrawing and adjusting the associated design to conform to the new guidelines. Therefore, designers currently use guidelines only as visual references or a way to document their thoughts and yet cannot benefit from the rich design information encapsulated by them.

Hence, using guidelines as the input conditions to our model can not only provide a high-level yet precise control to users, but can also overcome a practical challenge in layout design workflows across different domains. We consider the guideline condition as a type of \textit{parametric conditions}, borrowing the definition from parametric design in the fields of computational design and computer-aided design \cite{parametric}. In parametric design, the design artifacts are defined using algorithms or procedures with parameters as inputs, and therefore users can easily make design changes by manipulating the parameters. Our work enables an intuitive way to instantiate parametric models for layouts by drawing guidelines, as opposed to manually setting the relationships and heuristics in traditional parametric design tools.

We introduce \textit{PLay}, \textbf{P}arametrically Conditioned \textbf{Lay}out Generation using Guidelines, a two-stage model following the idea of latent diffusion models (LDMs) proposed by \cite{ldm}. Compared to \cite{ldm}, the main purpose of our first-stage model is to convert the layout from the discrete vector graphic space to the continuous latent space instead of compressing information in original data samples. With this continuous latent representation, the second-stage diffusion model can iteratively refine the results similar to image-based diffusion models. PLay can generate layout designs conditioned on guidelines across three different datasets with significantly improved quality. We measure the quality by computing the Fréchet distances using layouts rendered in both image and vector graphic domains. We further evaluate the results by conducting user study with professional designers.


Using guidelines as a way to express both high-level and low-level ideas, PLay also enables several interactive and controllable ways for users to create desired layouts: 1) generating and editing layouts by dragging, adding, and removing guidelines, similar to drawing a sketch, 2) generating variations from an existing layout with different levels of similarity controlled by guidelines, and 3) layout inpainting. Benefiting from our guideline sampling schema during training, we further reduce the effort for users to draw guidelines, as PLay only requires users to specify the guidelines they think are important. Moreover, in practice, guideline templates are often reused across projects and can be extracted from existing designs, providing a wide range of low-effort use cases for design generation and editing using PLay. We envision PLay to improve layout design workflows in various domains, enable different types of conditions and interactions, and potentially solve more general vector graphic problems.

In summary, the main contributions of this paper are:
\begin{itemize}
\item We develop a latent diffusion model in the vector graphic domain for layout generation, achieving a better performance than prior work in multiple metrics with large margins.
\item We introduce guidelines as input conditions for the latent diffusion model, making the generation process parametrically controllable and interactive.
\item We provide a variety of new ways to generate and edit layouts, including guideline editing, layout inpainting, and generating variations from existing layouts.
\item We propose FID and FD-VG as metrics and conduct user study to evaluate the quality of generated layouts.
\end{itemize}

\section{Related Work}
\subsection{Layout Generation}
Several prior works have studied generative models for layouts in the vector graphic domain. LayoutGAN \cite{layoutgan} is among the earliest ones---it uses self-attention layers as the generator and argues that the discriminator in the image domain can better evaluate the spatial quality, such as alignments of the elements. LayoutVAE \cite{layoutvae}, on the other hand, works purely in the vector graphic domain. It aggregates the information across the elements using Long Short-Term Memory (LSTM) \cite{lstm} and trains variational autoencoders (VAEs) \cite{vae} for generation. LayoutMCL \cite{diverse} and LayoutTransformer \cite{layouttransformer} generate layout boxes auto-regressively. LayoutMCL uses a CNN+RNN structure with multi-choice prediction and winner-takes-all loss, whereas LayoutTransformer employs a Transformer decoder to output each box attribute as an individual token. A recent work, VTN \cite{vtn}, maintains the VAE architecture in LayoutVAE but replaces the encoder and decoder by Transformers \cite{transformer}. The results from VTN shows that with Transformers, the model can learn the proper layout arrangements without mapping results to the image domain. Our latent diffusion model in PLay can also be seen as a way to learn a better latent prior than VTN.

Various types of conditions for conditional layout generation have also been studied. For example, \cite{attribute} adds element attributes such as area and aspect ratio as conditions, and \cite{neuraldesign} uses graph neural networks to constrain inter-element relationships. BLT \cite{blt} is a conditional model that allows users to control the properties of each element using a BERT-based approach with a customized masking schema. However, it cannot be directly applied to our scenario as the guidelines are not part of the box attributes. In the image domain, House-GAN \cite{house} and House-GAN++ \cite{plusplus} also explore graph conditions for layout generation. However, the shared issues of these approaches are 1) users have to tediously assign the properties of relationships between each of the elements, and 2) the allowed number of elements and relationships for conditions are often too small for a complex layout with more than 20 objects. As discussed in \ref{intro}, the guideline conditions in PLay can overcome these issues, since one guideline can align multiple elements, and multiple guidelines can define complex rules. 

Guidelines can also be seen as space partitioning in 3D. For example, \cite{zonegraphs} partitions the 3D space to search for potential CAD modeling sequences, and \cite{building} uses the partitioned space as one of the input conditions for 3D volume generation. Both works use partitions directly as the design or search space, meaning that if the given partitions are not good enough, the quality of the results will be affected. In contrast, PLay can flexibly attend to various guidelines through cross-attention and generate elements that do not follow some guidelines if needed.

In addition, some recent works incorporate other modalities such as images into layout generation: CanvasVAE \cite{canvasvae} uses the image information for the elements in the layout generation process, CGL-GAN \cite{cglgan} and ICVT \cite{geometry} generates layouts that fit the given images, and LayoutDETR \cite{layoutdetr} leverages DETR to encode both background and element images. \cite{coarsefine} recently introduced hierarchical decoding to VTN, where the first-stage decoder generates the regions and the second-stage decoder generates the elements in a region. The limitation of this two-stage decoder is that it cannot generate layouts with structure depths larger than two, which are common in realistic UI layouts. For example, in CLAY, the max depth of a layout hierarchy is $10$. CanvasVAE has a two-stage encoder-decoder architecture inspired by DeepSVG \cite{deepsvg}, with the first-stage for the individual elements and the second-stage for the layout. In PLay, the element stage is not needed since each element in our datasets has a fixed-length representation composed of its class and bounding box coordinates only. However, it is worth experimenting with the element-level encoding stage for PLay in the future to solve more complex tasks, such as CAD layout generation, where each element can be any shapes or curves.



\subsection{Diffusion Models}

Recent advances in Diffusion Models (DMs) \cite{diffusion} have shown promising results in image generation \cite{score, ddpm, cg}, where classifier guidance \cite{cg} and classifier-free guidance (CFG) \cite{cfg} methods not only improve the generation quality but also enable the possibility of developing conditional diffusion models. Most recent works for text-to-image generation use CFG, including GLIDE \cite{glide}, DALL·E2 \cite{dalle2}, Imagen \cite{imagen}, and LDMs \cite{ldm}. In addition to text, LDMs explore other conditions such as bounding boxes and semantic maps. We also adopt CFG for the guideline conditions in PLay, as it does not require an extra classifier and leads to the generation of better results.

Applying diffusion models outside of the image domain has also drawn attention from researchers, such as 3D-model \cite{magic3d}, video \cite{imagenvideo}, and music \cite{music} generation. \cite{music} converts discrete melody tokens into a continuous latent space, and trains the diffusion model in the latent space. This work inspires us to convert the discrete layout elements, composed by concatenating different types of tokens including their class and coordinates, to a continuous domain for the diffusion process. A concurrent work, \cite{language}, applies the same idea for language generation. We also follow LDMs \cite{ldm} to add a small KL-penalty to ensure high layout reconstruction quality and avoid arbitrarily large variance in the latent space.

Several recent works explore methods to further control the diffusion process. For instance, Prompt-to-Prompt \cite{prompt2prompt} fixes the cross-attention maps to preserve scene compositions for a new text prompt; SDEdit \cite{sdedit} injects the stroke-based conditions by adding noise to them and denoising them back to a real image. The level of added noise becomes a parameter to control the balance between image realism and faithfulness to the input drawing. Compared to these works, PLay can provide more explicit control to generate variations of previously generated layout, by extracting and editing the guidelines (\ref{variationbygls}).

\section{Layout Design}
\subsection{Datasets}
We experiment PLay with three publicly available datasets for two different domains: UI and document layouts.
\begin{itemize}
\item \textbf{CLAY} \cite{clay} contains about 50K UI layouts with 24 classes.
\item \textbf{RICO-Semantic} \cite{ricosemantic} contains about 43K UI layouts with 13 classes previously used in VTN.
\item \textbf{PublayNet} \cite{publaynet} contains about 330K document layouts with 5 classes.
\end{itemize}

The layouts in CLAY are more complex and representative of real UI designs compared to RICO-Semantic. Although both of them are extracted and processed from RICO \cite{rico}, CLAY tries to fix some annotation errors and mismatches between the screenshots and view hierarchies and introduces new label systems, whereas RICO-Semantic adds semantic annotations for RICO. We suspect that either RICO-Semantics filtered some complex patterns during post-processing, or the chosen 13 classes from the 25 original classes largely reduced the complexity.

To verify that PLay can be applied to other layout domains, we also train it on PublayNet to generate document layouts. Compared to CLAY, both RICO-Semantic and PublayNet have fewer design variations, and their layouts are overall simpler and have more repetitive patterns. For example, the complexity of each dataset is reflected by its average number of elements in each layout: RICO-Semantic=8.79; PublayNet=7.90; CLAY=19.62. Therefore, while we evaluate PLay over all the three datasets, we particularly conduct in-depth evaluation and analysis of our model on CLAY with qualitative and quantitative results. See the appendix for more statistics and examples of the datasets.


\subsection{Layout and Guidelines}
\label{layout_and_gl}

The shared data format across the three datasets for a layout is a sequence of elements: $E=\{e^1, e^2, ..., e^N\}$, where $e^n = \{{c}^n, x_{min}^n, y_{min}^n, x_{max}^n, y_{max}^n\}$ and $1\leq{n}\leq{N}$. The class of an element, ${c}^n$, is represented as a one-hot vector. Following the prior works \cite{skexgen, layouttransformer, solidgen}, we also found that discrete coordinate values work better empirically and set the dimension of each layout with ${width=36}$ and ${height} = 64$. The class and coordinates are then concatenated as a single vector, and therefore $E \in \{0, 1\}^{N \times D}$. We fix the maximum number of elements per layout: $N=128$, and the layout with fewer elements are padded to the same size, which result in fixed $N$ and $D$ for all layouts. This format can potentially be extended to represent general vector graphic elements by allowing more shape parameters, instead of 4 coordinates, and additional properties such as fill and stroke colors. 

The guidelines of a layout is represented as:  $G = \{g^1, g^2, ..., g^M\}$, where $g^m = \{{a}^m, p^m\}$ and $1\leq{m}\leq{M}$. Each guideline is composed by its axis $a^{m}$, i.e., horizontal versus vertical, and the coordinate position $p^{m}$. We fix the maximum number of guidelines for each layout: $M=128$, and thus the maximum number of guidelines in each axis is $M \slash 2$. The representation of layouts involving fewer guidelines is padded. To create the layout-guidelines pairs for training, for each layout $E$ in the datasets, we can intuitively obtain the full guidelines $G_{full}$ by extending all the bounding box edges of each element and removing the duplicated values. For the model to learn how to synthesize valid details beyond given all guidelines, we also have three different sampling methods to create random subsets of $G_{full}$ during training. More details will be discussed in \ref{gl_methods}.

\subsection{Metrics}
There are no universally established metrics to evaluate layout generation. Prior works such as \cite{vtn, layoutgan} compute several features in both real and generated layouts such as IoU, overlap, and alignment \cite{attribute}; \cite{read} proposes DocSim, measuring the feature-wise layout similarity; and \cite{canvasvae} computes the feature-wise distribution differences. Instead of using feature-based methods, we follow the common practice of evaluating Generative Adversarial Networks (GANs), including several prior layout generation works \cite{house, plusplus, neuraldesign}, to measure the Fréchet Distance (FD) between two distributions from latent space. To capture various aspects of layout generation, we compute FD in two ways with sample size $s=1024$: 

\begin{itemize}
\item \textbf{FID} \cite{fid}: we render the layouts into images with the same aspect ratio and add paddings if the elements are not fully using the screen space. We then feed the images into the pre-trained Inception \cite{inception} model to get the activation vectors, and compute the FD between the real and generated groups of layouts
\item \textbf{FD-VG}: we train a Transformer-based auto-encoder in the vector graphic domain, use its encoder to encode generated and real layouts, and compute the FD between them.
\end{itemize}

\textbf{G-Usage}:
We also evaluate if the generated results satisfy the guideline conditions by computing the G (guideline)-Usage. We first extract the guidelines $G^\ast$ from a generated layout, then we calculate the intersection, $G^{inter}$, for $G^\ast$ and the given guidelines, $G$. Finally, we obtain the G-Usage as $|G^{inter}| \slash |G|$ and take the average across the generated samples. Note that G-Usage does not equal to IoU, since it is acceptable to have guidelines in $G^\ast$ that are not part of $G$.

\textbf{User Study}:
We also conduct user study (\ref{user_test}) with professional designers and the results are aligned with the FID and FD-VG metrics.

\begin{figure*}[ht]
\begin{center}
\centerline{\includegraphics[width=0.72\textwidth]{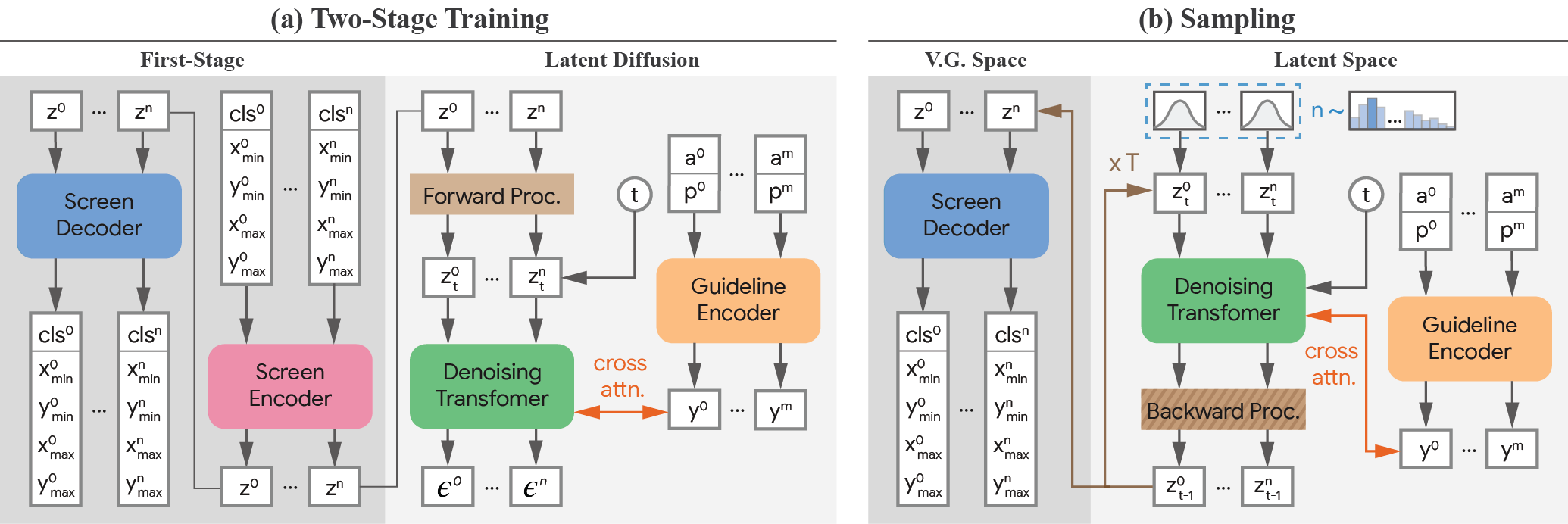}}
\caption{Model architecture. After training the first-stage model, we can use it to encode layouts to the latent space for training the latent diffusion model. During sampling, it can decode the generated latent representations back to layouts.}
\label{architecture}
\end{center}
\end{figure*}

\begin{figure*}[ht]
\begin{center}
\centerline{\includegraphics[width=0.93\textwidth]{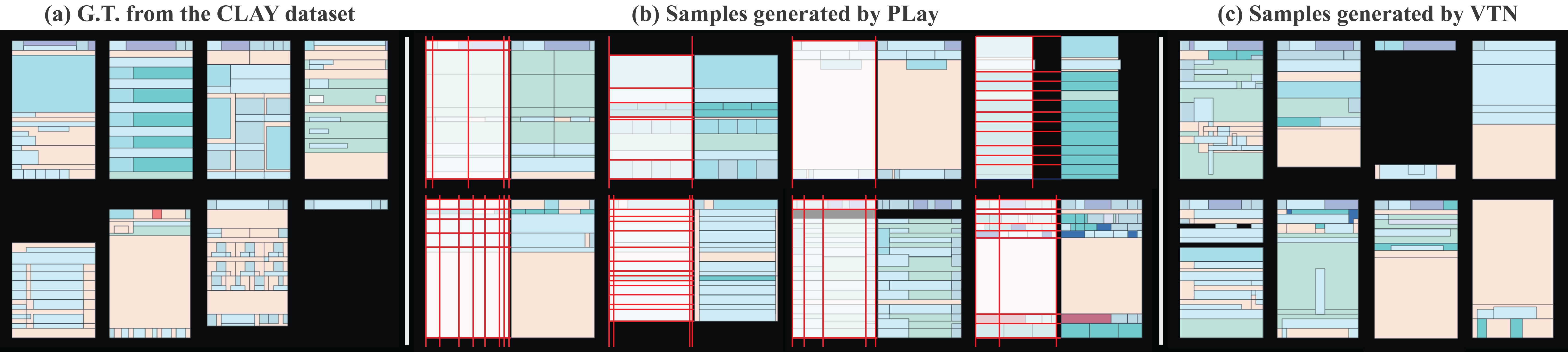}}
\caption{Qualitative results. We can observe that PLay generates reasonable results on complex examples with many elements, whereas VTN often struggles in such cases.}
\label{qualitative}
\end{center}
\end{figure*}


\section{Architecture}

Following the image-based latent diffusion model \cite{ldm}, We formulate PLay as a two-stage model, where the first-stage model learns to map layouts from a vector graphic space to a latent space, and then the conditional latent diffusion model learns to generate layouts in latent space conditioned on guidelines given by the users. Figure~\ref{architecture} illustrates the overview of our model.

\subsection{First-Stage Model}
To map the layout $E \in \mathbb{R}^{N \times D}$ to the latent representation $z \in \mathbb{R}^{N \times d}$, we use a Transformer similar to DETR \cite{detr}, with an encoder $\mathcal{E}(E)$ and a non-autoregressive decoder $\mathcal{D}(z)$ modified from DeepSVG \cite{deepsvg}. We also added a small KL-penalty to regularize the latent space while keeping the high reconstruction accuracy, which is especially critical in the vector graphic space. The reason is that unlike pixels, where some artifacts and noises are not noticeable, we only have at most 128 elements in a layout, so it will be obvious if any of them is decoded incorrectly, such as having a misaligned box or an unreasonable class. Following VTN, no positional embeddings are added to the encoder and decoder, as the coordinates already explicitly indicate the positions spatially. 

In image-based latent diffusion models (LDMs), the goal of the first-stage model is to map the input image to a lower dimension while keeping the same perceptual details. However, the first-stage model in PLay serves a different purpose: to learn a meaningful and continuous latent representation of the discrete vector graphic space. We will see in \ref{sub_base}, that the LDM cannot converge with naive mapping using MLP layers. In addition, we experiment with encoding the entire layout as a single vector using a transformer, but it also fails to achieve high reconstruction accuracy.

\subsection{Latent Diffusion Model}
When training the latent diffusion model, the encoded layout $z$ is first divided by the standard deviation ${std}$ of the first batch, as suggested by \cite{ldm}, and then the scaled $z$ is used as $z_0$ for the forward diffusion process to get $z_t; t=1...T$. For the denoise network $\epsilon_\theta(z_t, \tau_\psi(G), t)$, we use a Transformer encoder to replace the U-Net structure used in image-based DMs and predict the noise $\epsilon$. The discrete time step $t$ is encoded using Feature-wise Linear Modulation \cite{film} and injected into the Transformer encoder with a feature-wise affine layer. We encode the guidelines $G$ using another Transformer encoder $\tau_\psi$, and the encoded guidelines, $\tau_\psi(G)\in{\mathbb{R}^{M\times d}}$, are then fed to $\epsilon_\theta$ through cross-attention. The loss function can be formulated similar to general LDMs:

\begin{equation}
L := \\ \mathbb{E}_{\mathcal{E}(x), G, t, \epsilon \sim \mathcal{N}(0, 1)}\left[ \parallel \epsilon - \epsilon_\theta(z_t, \tau_\psi(G), t)\parallel^2 \right]   
\end{equation}

We train the LDM as a standard DDPM \cite{ddpm} with classifier-free guidance, where we randomly drop the guideline conditions with the probability $p_{drop}=0.1$.

\subsection{Sampling}
In sampling, we first either sample the number of elements $N$ from $p(N)$, which is the element count distribution of the dataset, or use $n$ assigned by the user. Then we initialize $z_T$ and denoise it to get $z_0$ with the given guideline conditions using DDPM and CFG, with $w=1.5$:
\begin{equation}
\begin{split}
    \hat{\epsilon}_\theta(z_t, \tau_\psi(G), t) &= (1+w)\epsilon_\theta(z_t,  \tau_\psi(G), t) \\ &- w \epsilon_\theta(z_t,  \tau_\psi(\phi), t) 
\end{split}
\end{equation}
We then rescale $z_0$ with the ${std}$ used in training, and decode it back to the vector graphic domain using the first-stage decoder: $E = \mathcal{D}(z)$.

\begin{table}[t]
\caption{Quantitative Results and ablation studies.}
\label{ablation}
\vskip 0.15in
\begin{center}
\begin{small}
\begin{tabular}{l|cc|ccl}
\toprule
Model & F.S. & CFG-W & FID & FD-VG & G-Usage\\
\midrule
VTN & $\times$ & $\times$ & 19.10 & 0.352 & n/a \\
C-VTN & $\times$ & $\times$ & 16.22 & 0.361 & 0.819 \\
\midrule
PLay & VAE & 1.25 & 12.63 &0.286 & 0.970\\
 & VAE & 1.50 & \textbf{10.59} & \textbf{0.245} & 0.964\\
 & VAE & 1.75 & 11.21 & 0.269 & 0.968\\
\cmidrule{2-6}
 & $\times$ & 1.5 & 166.7 & 4.577 & \textbf{0.992}$^*$\\
 & VAE & $\times$ & 14.80 & 0.375 & n/a\\
 & VTN & 1.50 & 14.35 & 0.311 & 0.835\\
 & VQVAE & 1.50 & 11.49 & 0.254 & 0.937\\
\bottomrule
\end{tabular}
\end{small}
\end{center}
\footnotesize{$^*$PLay without the first-stage model generates mostly random, unaligned boxes and therefore, has a high guideline usage.}
\vskip -0.1in
\end{table}

\begin{table}[t]
\caption{Comparisons across datasets.}
\label{datasets}
\vskip 0.15in
\begin{center}
\begin{small}
\begin{tabular}{l|lccc}
\toprule
Dataset & Model & FID & FD-VG & G-Usage\\
\midrule
CLAY & VTN & 19.10 & 0.352 & n/a \\
& PLay & \textbf{10.59} & \textbf{0.245} &\textbf{0.964}\\
\midrule
RICO- & VTN & 18.80 & 0.415 & n/a\\
Semantic & PLay & \textbf{13.00} & \textbf{0.320} & \textbf{0.944}\\
\midrule
PublayNet & VTN & 19.80 & 0.787 & n/a\\
 & PLay & \textbf{13.71} & \textbf{0.408} & \textbf{0.971}\\
\bottomrule
\end{tabular}
\end{small}
\end{center}
\vskip -0.1in
\end{table}

\begin{table}[t]
\caption{FID scores in groups of number of elements.}
\label{groups}
\vskip 0.15in
\begin{center}
\begin{small}
\begin{tabular}{l|c|ccccc}
\toprule
Model & All & 1-6 & 7-11 & 12-18 & 19-29 & 30+\\
\midrule
VTN & 19.10 & 11.83 & 16.12 & 19.89 & 25.76 & 30.15\\
PLay & \textbf{10.59} & \textbf{9.28} & \textbf{11.72} & \textbf{13.23} & \textbf{13.60} & \textbf{14.01}\\
\bottomrule
\end{tabular}
\end{small}
\end{center}
\end{table}

\begin{table}[t]
\caption{Guideline sampling methods.}
\label{sampling_methods}
\vskip 0.15in
\begin{center}
\begin{small}
\begin{tabular}{l|ccc}
\toprule
Method & FID & FD-VG & G-Usage\\
\midrule
All & 12.11 & \textbf{0.206} & 0.917\\
Uniform & 11.89 & 0.312 & 0.944\\
Weight-tiers & 11.70 & 0.297 & 0.957\\
Weighted & \textbf{10.59} & 0.245 &\textbf{0.964}\\
\bottomrule
\end{tabular}
\end{small}
\end{center}
\vskip -0.1in
\end{table}

\begin{figure}[ht]
\begin{center}
\centerline{\includegraphics[width=0.32\columnwidth]{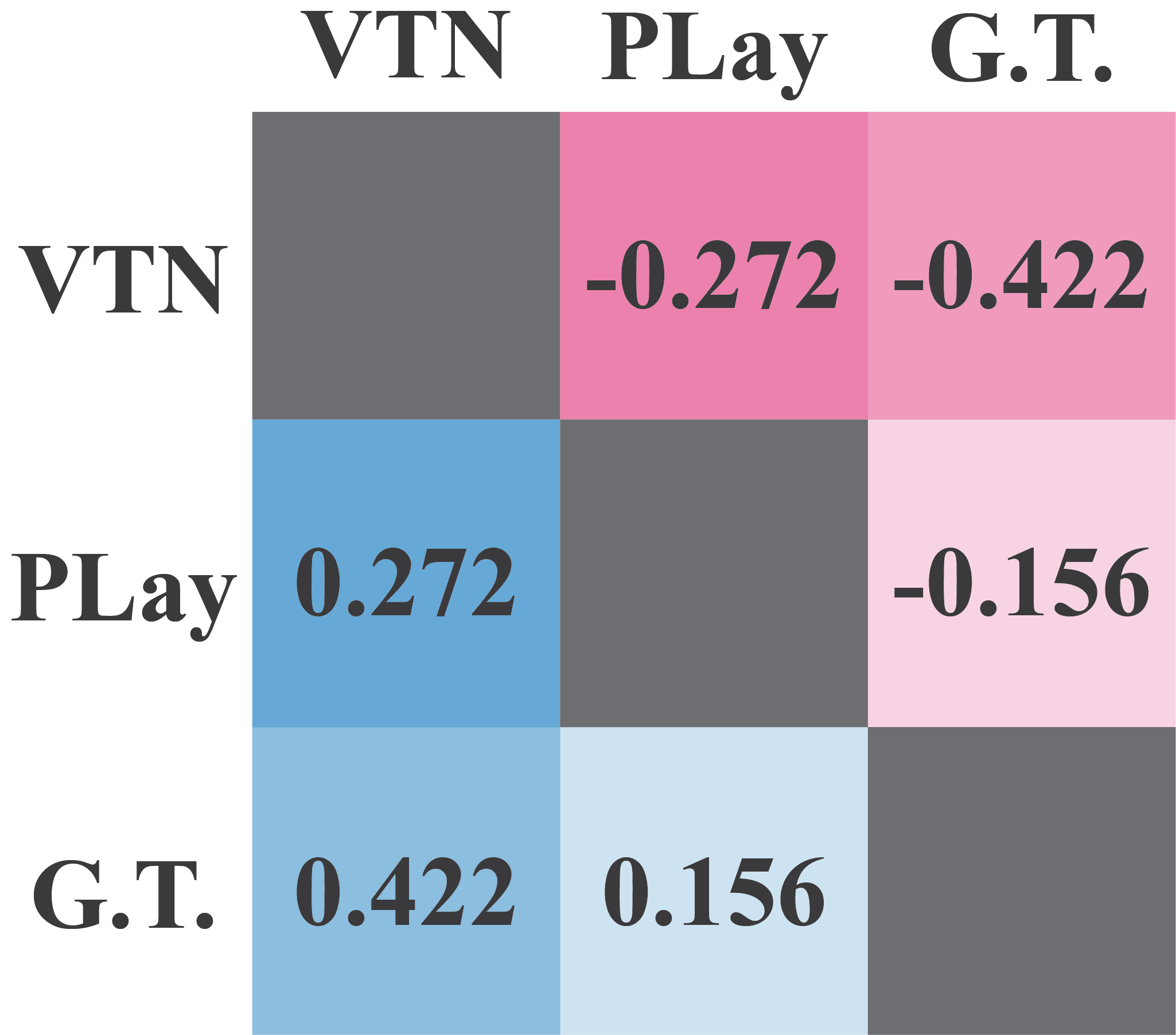}}
\caption{User study results. The score between each combination of the sample pools (GT-PLay, GT-VTN, and PLay-VTN) is calculated by subtracting the number of times that a pool is ranked worse than the other pool from the number of times that a pool is ranked better than the other one, normalized by the total number of comparisons between two pools. For example, if pool A was ranked better than pool B $50$ times and worse than pool B $30$ times, then the score for A would be $20/100 = 0.2$.}
\label{user}
\end{center}
\end{figure}

\section{Experiments}

\subsection{Baseline Comparison and Ablation Study}
\label{sub_base}
In this section, we show how PLay performs compared to the baseline and study the effects of different first-stage model choices and classifier-free guidance weights. We choose VTN \cite{vtn} as the baseline model for comparison because it is the most common framework of SoTA models for UI layout generation in the vector graphic space and is easily reproducible. It shares the same architecture with our first-stage model---a variational authoencoder (VAE). We also modify VTN to condition on guidelines, called C-VTN to ensure a fair comparison against PLay. In Table~\ref{ablation}, we first try to train a diffusion model without the first-stage model. With simple MLP layers to encode the class and coordinates, the model fails to converge. We then add VAE as the first-stage model, and discover the results of latent diffusion training outperform the results of the baseline. Adding the classifier-free guidance further improves the results, with the optimal weight $w=1.5$ and latent dimension $d=8$ for our experiments. 

We also try to use VQVAE as the first-stage model, but it falls short on G-Usage while having comparable numbers in FID and FD-VG. The potential reason can be that the learned, frozen codebook is less flexible to exactly match the guideline conditions since guidelines are not involved in the first-stage training. Additionally, we try to use the trained VTN as the first-stage model, as it is also a VAE with KL loss weight $\beta=1.0$. We achieve worse but still reasonable results, which can be explained by the low reconstruction accuracy using high $\beta$ values. The reconstruction accuracy plays a crucial role for PLay, and it is not required to use a very small $\beta$ value, e.g., $1e^{-6}$ in \cite{ldm}, as long as the reconstruction accuracy is high enough. See the appendix \ref{fs_appendix} for the complete table of first-stage model choices.

PLay also outperforms prior work in all metrics across three datasets: CLAY, RICO-Semantic, and PublayNet (Table~\ref{datasets}). Moreover, for this experiment, we use the same model architecture and hyper-parameters for all datasets. The only required change is the input dimension, which demonstrates PLay's ability to generalize to different layout domains.

We further divide the generated layouts into groups by the number of elements in each layout. We discover that PLay achieves much better FID scores (Table \ref{groups}) and is qualitatively better (Figure \ref{qualitative}) than VTN in groups with a large number of elements. This shows that the advantage of PLay over the baseline is more significant when a layout is complex and involves a large number of elements.

\subsection{Guideline Sampling Methods}
\label{gl_methods}
Users usually prefer to specify only the guidelines that can represent the main ideas instead of drawing guidelines for every element, because at this stage they have not come up with all the details for the design yet and want to see the potential options. Therefore, we randomly sample a subset of guidelines for every example during training. In this way, the model can learn to follow the given guidelines as the main guidance and create extra details that are not covered by the given guidelines. For example, a UI designer might start with a simple idea, such as creating a two-column layout. In this case, only five guidelines, similar to the top-left example in Figure \ref{teaser}, would need to be drawn. In addition, we can easily extend our approach to enable local or hierarchical guidelines, which gives designers more fine-grained control of guidelines in some situations.

We investigate three guideline sampling methods during training: uniform, weighted, and weight-tiers. Uniform means we uniformly sample a subset of the guidelines. For weighted sampling, we compute the weight for each guideline by summing the length of element edges that overlap with it. A guideline is deemed more important thus has a higher weight if total length of the overlapped edges is longer. For weight-tiers, we further bin the guidelines with different ranges of weight into groups and sample the groups as a whole. We find that weighted sampling achieves the best FID and G-Usage (Table ~\ref{sampling_methods}). Including all guidelines has the best FD-VG score while the G-Usage suffers since it becomes a more difficult task to fit all guidelines. Importantly, the model that is trained with all guidelines can only strictly follow the given guidelines and is incapable of creating elements beyond the guidelines when detailed guidelines are not given by the user, which is often the case in layout design.

\subsection{User Study}
\label{user_test}
We conduct a user study (Figure \ref{user}) to further evaluate the quality of generated layouts and whether the FID and FD-VG metrics align with human evaluation conducted by professional designers. We follow the method used in \cite{house} and \cite{building} and invite $28$ designers with user interface design expertise. The range of the scores is $[-1, 1]$, where $1$ means a model winning all comparisons, $-1$ means losing all, and $0$ means drawing all. Our result shows that although designers still prefer the ground truth samples over both VTN and PLay, the margin between the ground truth and PLay is small, and PLay wins over VTN by a large margin. In other words, professional designers consider PLay generating more realistic layouts than prior work and often favor PLay over the ground truth samples.

\subsection{Conditional Generation and Guideline Editing}
Designers commonly modify existing layouts and build upon their earlier designs. Therefore, we develop four ways for users to interact with the model using guidelines as input conditions. The goal is to create a fast and controllable workflow for users to iterate and refine generated layouts.

\subsubsection{Generating Variations from Given Design}
\label{variationbygls}
\begin{figure}[ht]
\begin{center}
\centerline{\includegraphics[width=0.9\columnwidth]{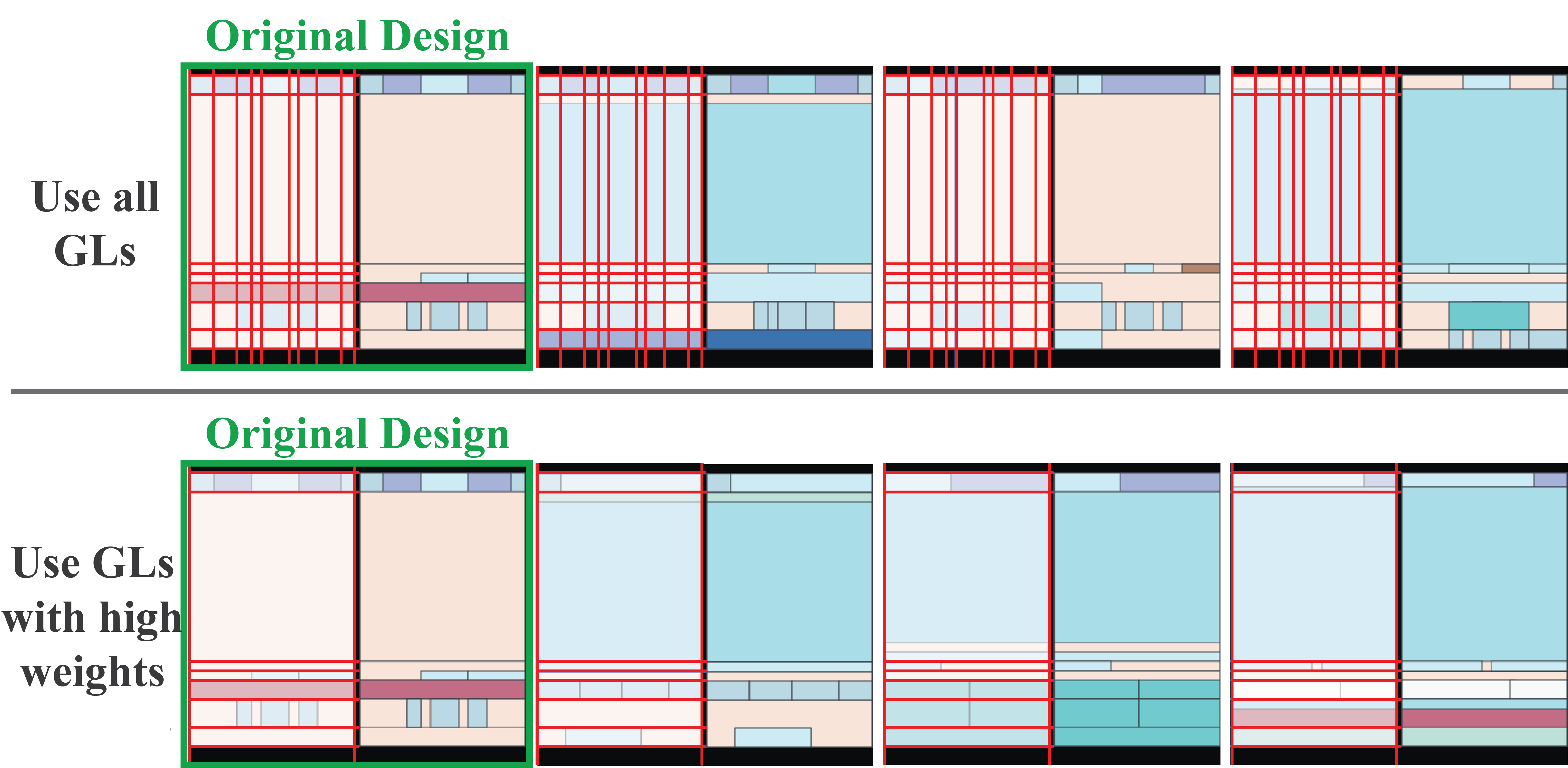}}
\caption{Generating variations from a given design using extracted guidelines. The results in the top row use all the extracted guidelines and therefore are similar to the original one in detail. The results in the bottom row have richer variety but are less similar with the original since they are less constrained.}
\label{variations}
\end{center}
\end{figure}

As discussed in Section~\ref{intro}, guidelines can represent design intentions, rules and element arrangement patterns. Therefore, they can be seen as a high-level abstraction or skeleton of an existing design, similar to sketches. Based on this observation, we develop a method for PLay to create variations from an existing layout, by first extracting the guidelines from the given layout and then using the extracted guidelines as conditions to generate more layouts. In Figure~\ref{variations}, all the generated results share the same high-level idea with the original one. By using different numbers of guidelines, we can control the levels of similarity to the given design. This variation to similarity trade-off is close to SDEdit \cite{sdedit}, but our method can give the user explicit and visual control over the level of similarity and where it needs to be similar in the layouts. Note that the number of elements is fixed in this experiment.

\begin{figure}[ht]
\begin{center}
\centerline{\includegraphics[width=0.9\columnwidth]{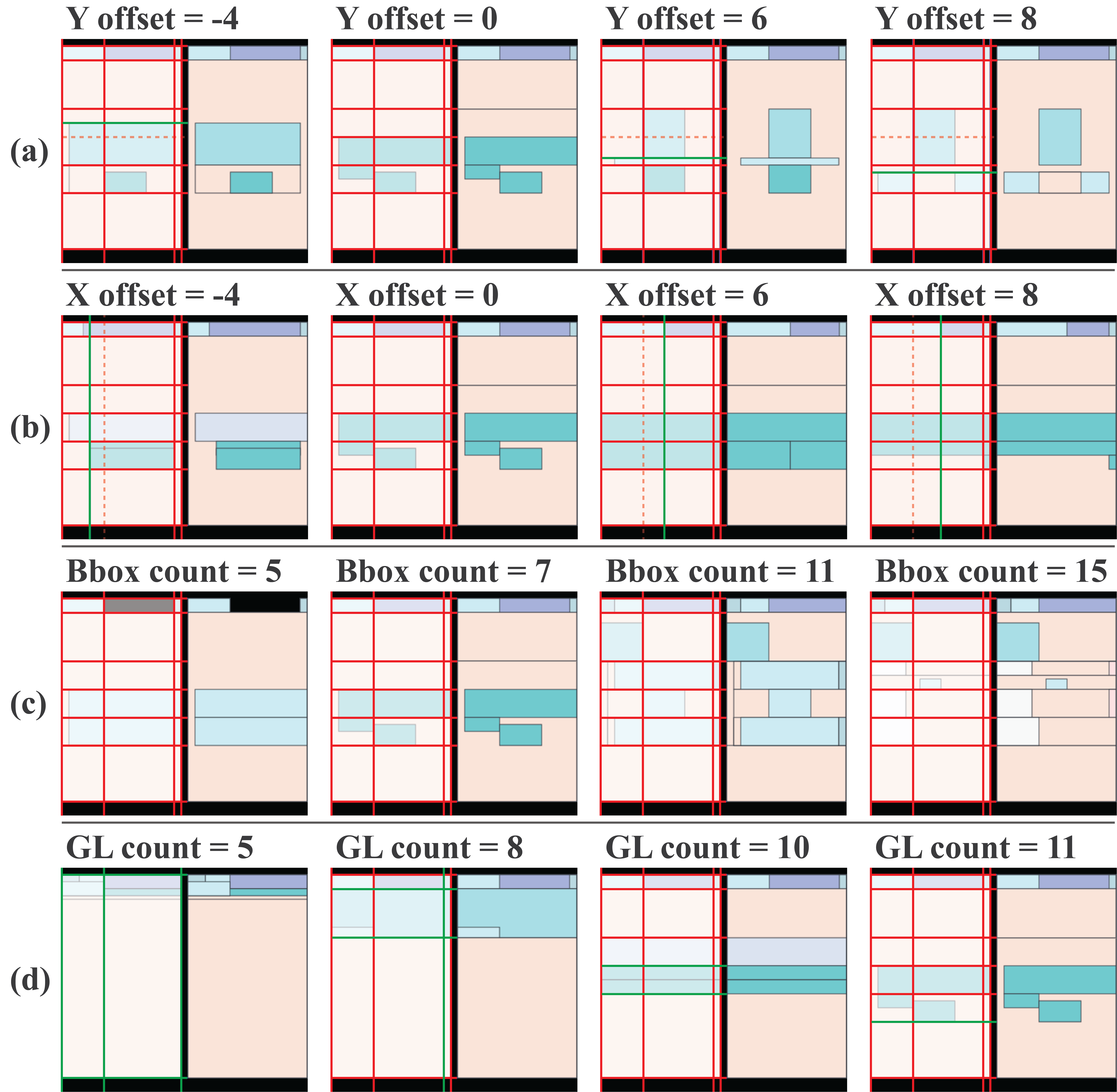}}
\caption{Layout editing. Row (a) and (b): drag guidelines along the x and y axes. Row (c): change the number of elements while keeping the same guidelines. Row (d): gradually draw new guidelines.}
\label{gl_editing}
\end{center}
\end{figure}

\subsubsection{Moving Guidelines}
Besides generating variations, users also need to further edit the generated results. One of the intuitive ways to edit the design is to allow the user to adjust a guideline by dragging it to a new position, expecting the elements around this guideline to be adjusted automatically while holding the rest parts of the layout intact. We achieve this by recording the noise values in diffusion steps, and reuse them to generate a new layout with the edited guidelines and the same number of elements. In Figure~\ref{gl_editing}a and~\ref{gl_editing}b, we show the results of changing guideline positions along the X and Y axes.

\subsubsection{Adding and Removing Guidelines}
Similar to moving guidelines, by fixing the number of elements and reusing the noise values, we can enable users keep adding or removing guidelines and inspect how they affect the layout arrangements. This feature introduces a new experience for layout design similar to \cite{igan} and \cite{gaugan}, where layouts are being generated simultaneously after each of the guidelines is drawn (Figure~\ref{gl_editing}d).

\subsubsection{Changing the Number of Elements}
During sampling, the number of elements to have in a layout can be either sampled from the distribution learned from the dataset or assigned by the user, which becomes another way to generate variations. Users can specify a different number of elements to examine how the generated elements fit into the given conditions (Figure~\ref{gl_editing}c).

\subsection{Layout Inpainting}
\begin{figure}[ht]
\begin{center}
\centerline{\includegraphics[width=0.95\columnwidth]{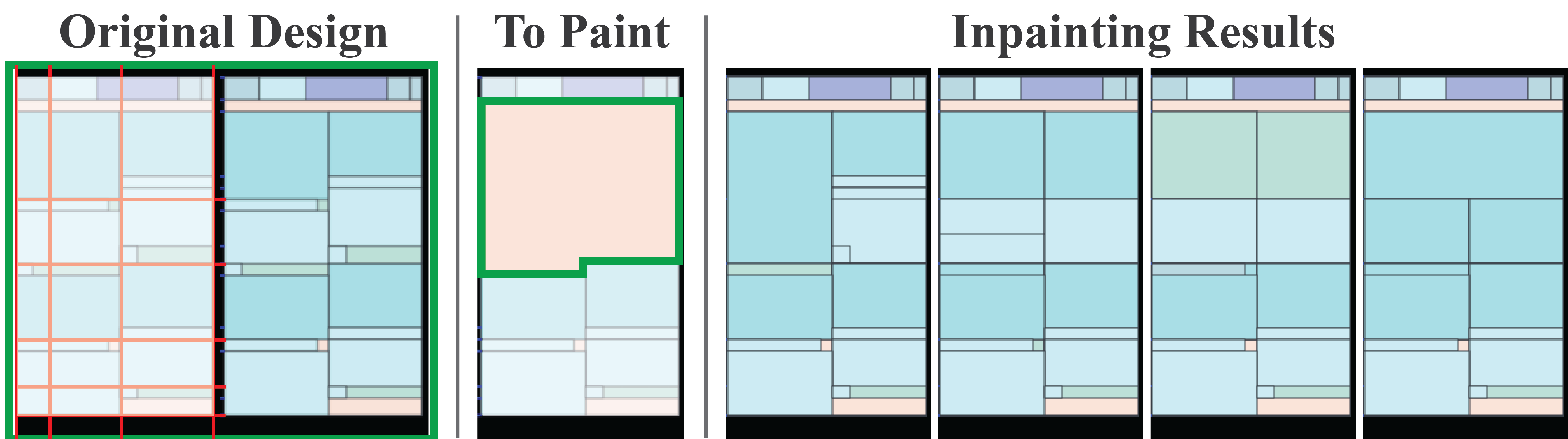}}
\caption{Layout inpainting. In this example, the inpainting results have better element alignments in the cropped area compared to the original design because of the guideline conditions.}
\label{inpainting}
\end{center}
\end{figure}

We report preliminary results for layout inpainting. In Figure~\ref{inpainting}, PLay generates new elements in the cropped area, and with the edited guidelines, the results have better element alignments in the cropped area compared to the original design. We adopt the inpainting method used in image-based diffusion models \cite{palette}. In this experiment, we simply match the number and sequence indices of the newly painted elements with the masked elements, which imposes a strong constraint and leads to results that are similar to the original design. Further study is needed to create a more flexible way to inpaint new elements, such as using an auto-regressive decoder to decide where to insert them.

\subsection{Failure Cases}
\begin{figure}[ht]
\begin{center}
\centerline{\includegraphics[width=0.9\columnwidth]{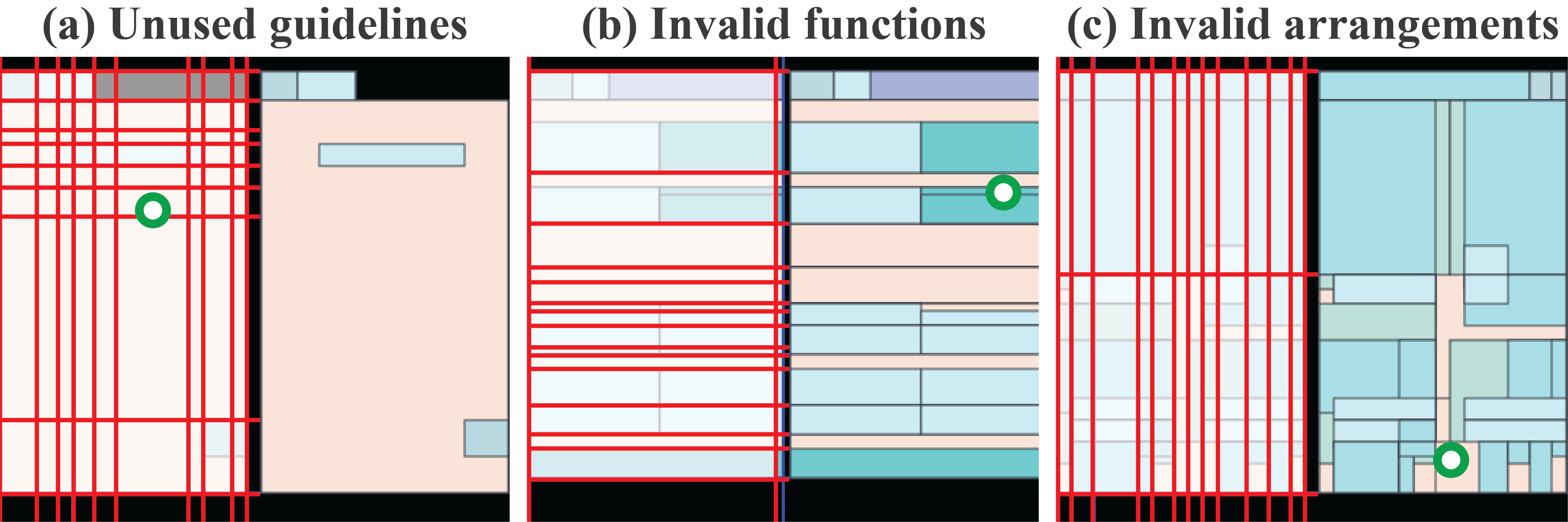}}
\caption{Failure cases. We can automatically compute if case (a) happens in a generated layout, but case (b) and (c) are more subjective and require designer's evaluation.}
\label{failure}
\end{center}
\end{figure}

We observe three common failure modes in samples generated by PLay. The first one is unused guidelines, and it can often be found when the number of elements is much lower than the number of guidelines, such as Figure~\ref{failure}a. The second type is invalid functions. For example, at the green dot in Figure~\ref{failure}b, it is unreasonable to have such a thin button for users to tap on. The last mode is invalid arrangements, which can be found when the number of elements are too large to be reasonably fit in a layout. Several future directions to improve these failure cases are: 1) training a generator to sample the number of elements based on guidelines instead of naively sampling from the data distribution, and 2) mitigating the imbalance of the number of elements distribution in the datasets.

\section{Conclusion}
We present PLay, a novel parametrically conditioned latent diffusion model for layout generation, and introduce guidelines, a widely used representation by designers, as our conditions. We achieve state-of-the-art results across three datasets on both qualitative and quantitative metrics, including FID, FD-VG, G-Usage, and via user study with designers. We also demonstrate different ways for users to control and interact with the generation process using guidelines, including guideline editing, inpainting, and generating variations from a given layout with user-controllable similarities.

\section*{Acknowledgements}

We thank the reviewers and area chair for providing constructive feedback. We also thank Ruiqi Gao for discussions and reviewing drafts of the paper.

\bibliography{play}
\bibliographystyle{icml2023}

\newpage
\appendix
\onecolumn
\section{Datasets}

\begin{figure}[ht]
\begin{center}
\centerline{\includegraphics[width=\columnwidth]{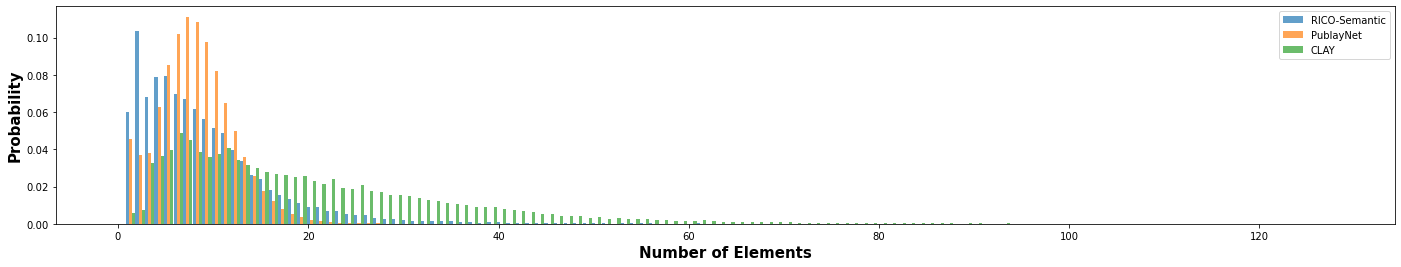}}
\caption{Number of elements distributions in CLAY, RICO-Semantic, and PublayNet.}
\label{noe_dist}
\end{center}
\vskip -0.2in
\end{figure}

\begin{figure}[ht]
\begin{center}
\centerline{\includegraphics[width=0.9\columnwidth]{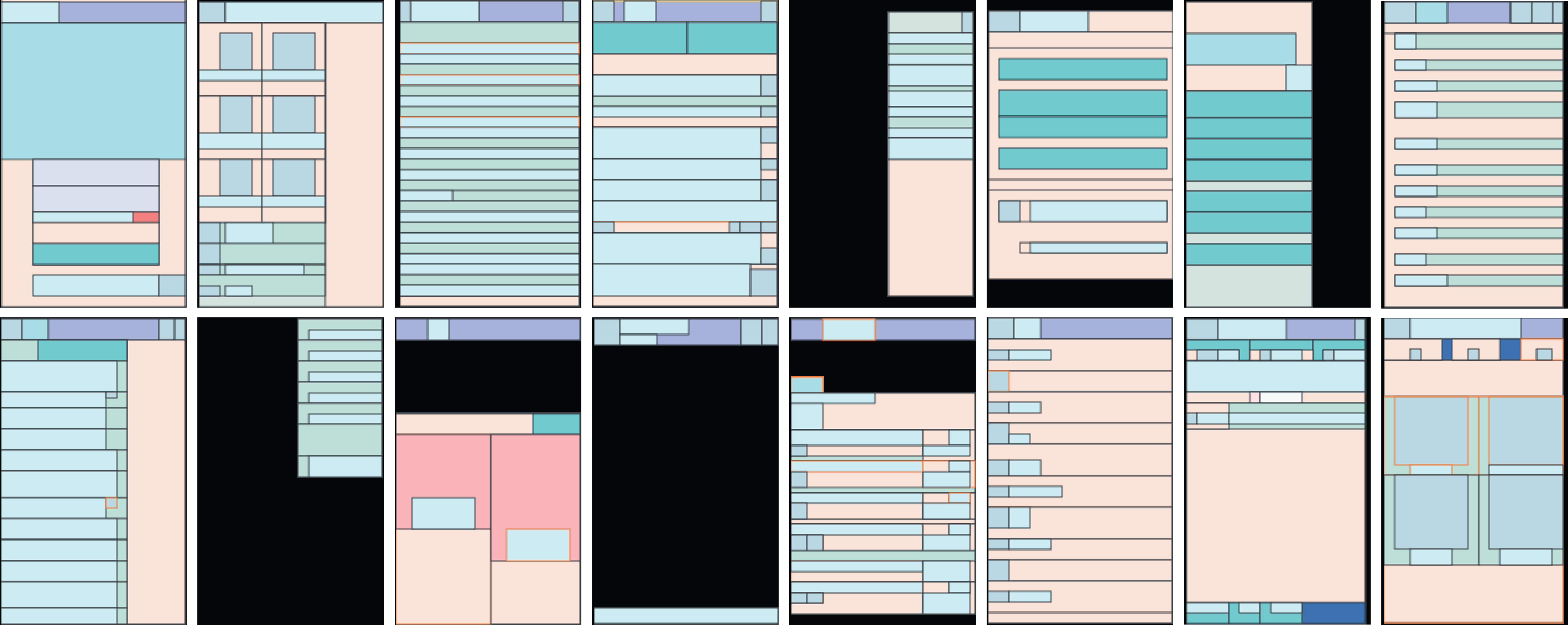}}
\caption{Examples from CLAY. Note that black is empty in all examples.}
\label{clay_exaples}
\end{center}
\vskip -0.2in
\end{figure}

\begin{figure}[ht]
\begin{center}
\centerline{\includegraphics[width=0.9\columnwidth]{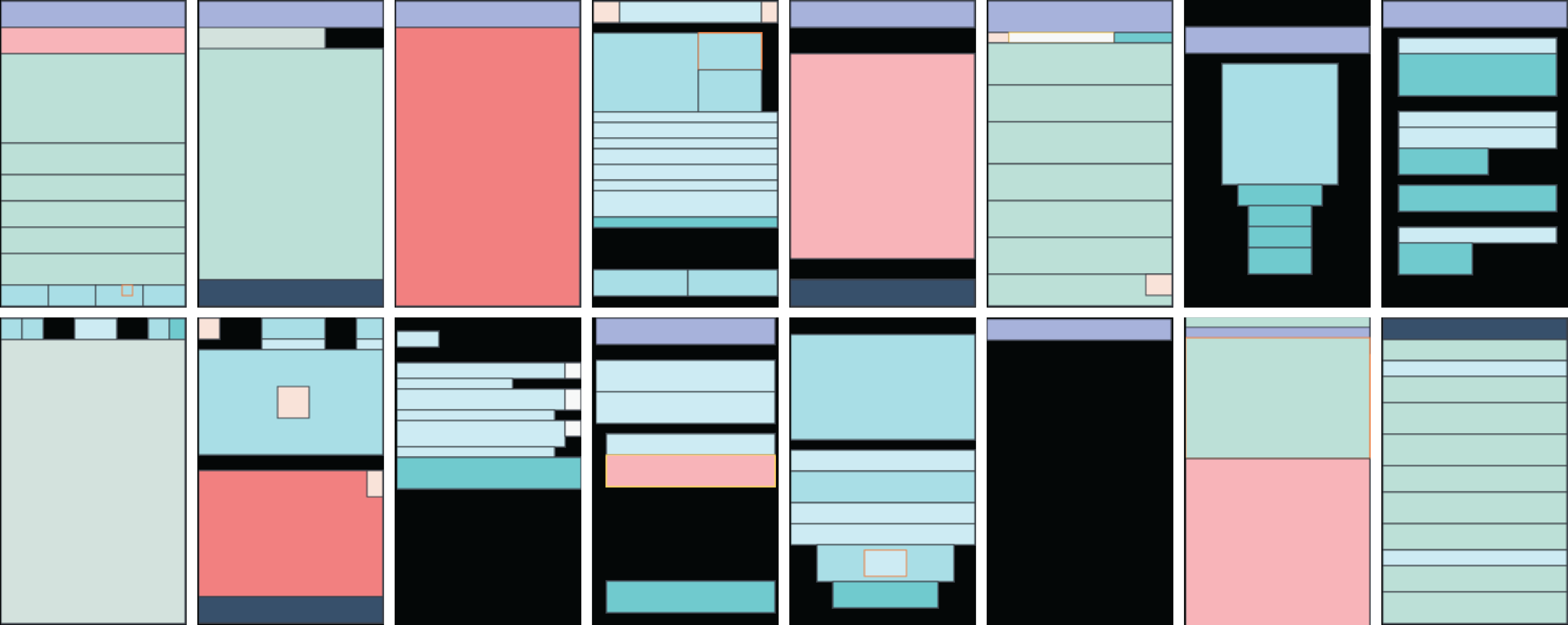}}
\caption{Examples from RICO-Semantic. Comparing the RICO-Semantic examples to the CLAY examples in Figure \ref{clay_exaples}, the number of elements in each layout is less; the average size of each element is larger; and the design patterns look simpler.}
\label{rico_exaples}
\end{center}
\vskip -0.2in
\end{figure}

\begin{figure}[ht]
\begin{center}
\centerline{\includegraphics[width=0.9\columnwidth]{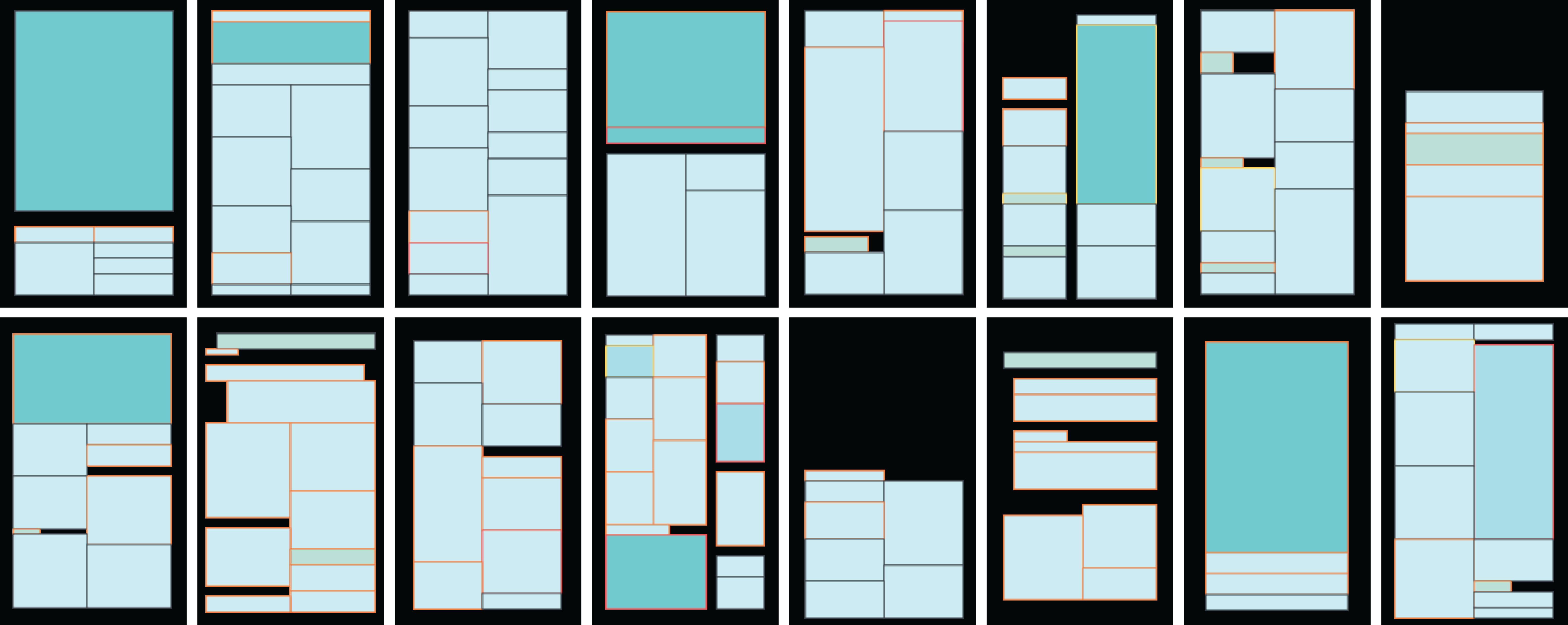}}
\caption{Examples from PublayNet.}
\label{pub_exaples}
\end{center}
\vskip -0.2in
\end{figure}

We can observe the difference in complexity of the three datasets from the number of elements distributions (Figure \ref{noe_dist}) and the visualization of examples (Figure  \ref{clay_exaples}, \ref{rico_exaples}, and \ref{pub_exaples}).

\section{Other Metrics}
In Table \ref{other_metrics}, We also computed the metrics used in VTN, including IoU, Overlap, and Alignment. PLay performs better than VTN on Overlap and Alignment, and its IoU is on par with VTN. However, compared to FID and FD-VG, these metrics do not align with the significant difference in our user study results. Moreover, PLay even outperforms the ground truth on Overlap and Alignment, which is not reflecting the fact that users still think the ground truth layouts are better than PLay. These metrics are also not commonly used for evaluating other design and creativity related generative models. Therefore, we choose not to use them as the main metrics.

\begin{table}[t]
\caption{Comparing PLay, VTN, and the ground truth layouts on IoU, Overlap, Alignment, and DocSim. The models are trained on the CLAY dataset.}
\label{other_metrics}
\vskip 0.15in
\begin{center}
\begin{small}
\begin{tabular}{l|cccc}
\toprule
Model & IoU & Overlap & Alignment & DocSim\\
\midrule
G.T. & \textbf{0.224} & 0.764 & 0.348 & n/a\\
VTN & 0.233 & 0.784 & 0.377 & 0.427\\
PLay & 0.235 & \textbf{0.754} & \textbf{0.302} & \textbf{0.513}\\
\bottomrule
\end{tabular}
\end{small}
\end{center}
\end{table}

\section{Implementation Details}
\begin{figure}[ht]
\begin{center}
\centerline{\includegraphics[width=0.9\columnwidth]{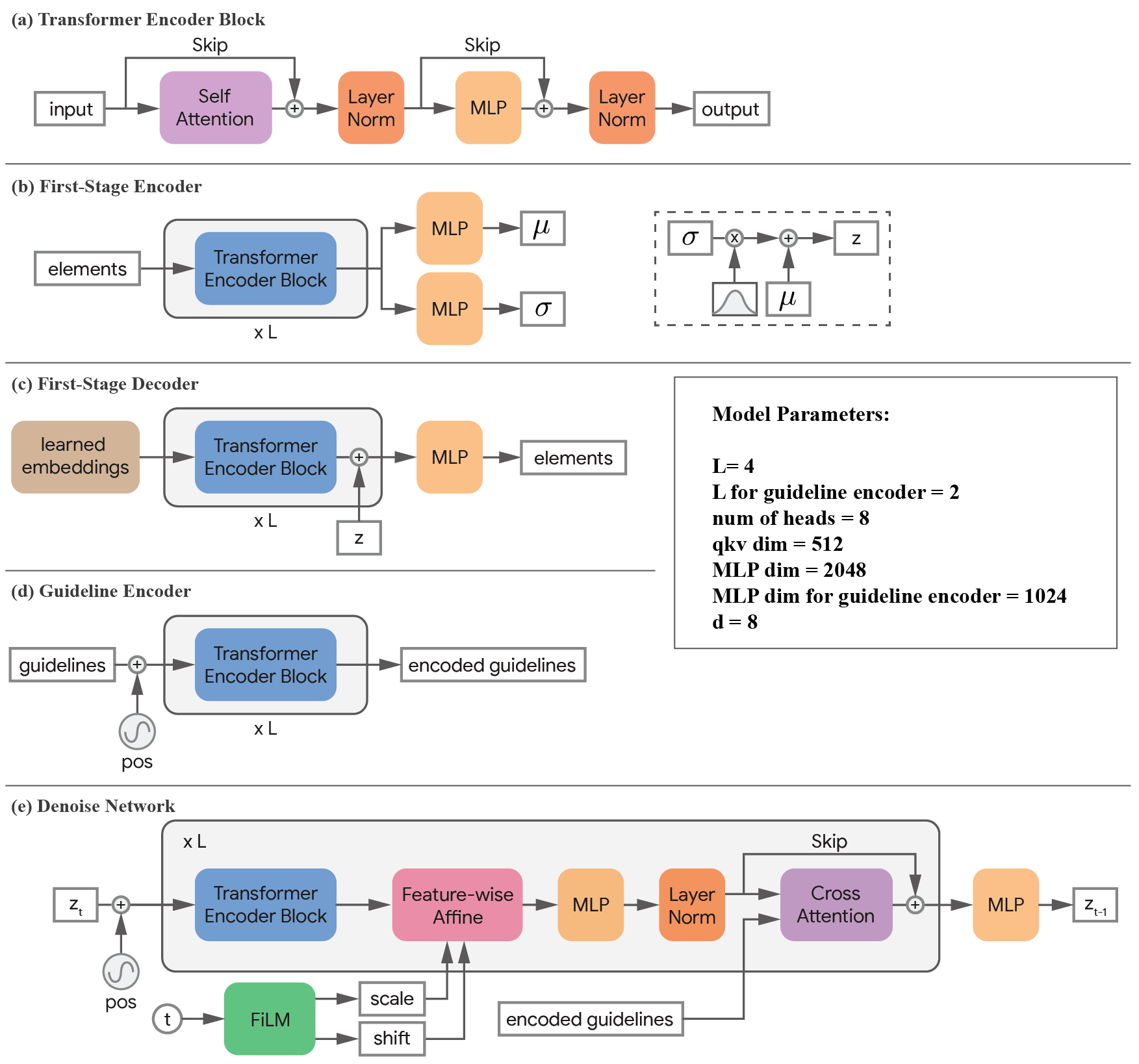}}
\caption{Model Components. We use the same Transformer Encoder (a) as the building block for all the components in PLay. The decoder (c) is similar to \cite{deepsvg}, using learned embeddings and adding $z$ in each layer. Note that positional embedding is needed in (e), since the encoded elements $z$ does not have explicit positional information (e.g., coordinates). The Feature-wise Linear Module is composed by MLP layers to map $t$ to scale and shift for the Feature-wise Affine layer.}
\label{architecture_details}
\end{center}
\vskip -0.2in
\end{figure}

The detail of each component in Figure \ref{architecture} can be found in Figure \ref{architecture_details}.

We implemented the proposed architecture in \textit{JAX} and \textit{Flax}. We use ADAM optimizer ($b_{1}=0.9$,  $b_{2}=0.98$) with 500k steps and a batch size of 128. The learning rate is $0.001$ with linear warming up to 8k steps. The model is trained using 8 Google Cloud TPU v4 cores for 47 hours.

\section{User Study Details}
\begin{figure}[ht]
\begin{center}
\centerline{\includegraphics[width=0.6\columnwidth]{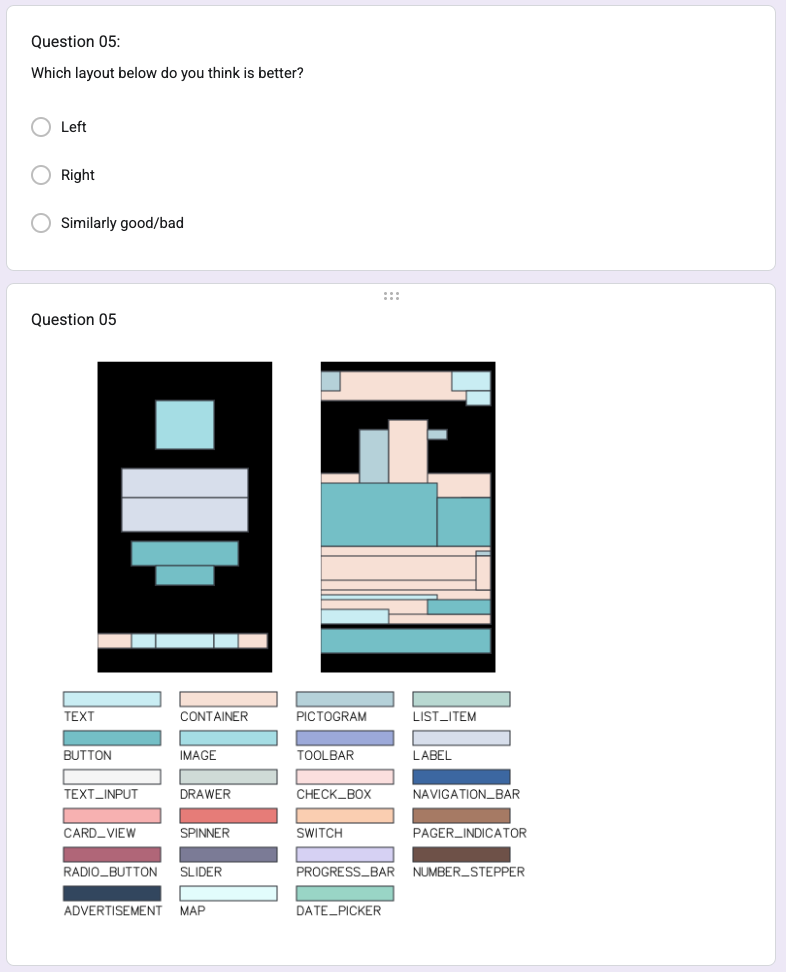}}
\caption{An example question in the user study.}
\label{user_test_q}
\end{center}
\vskip -0.2in
\end{figure}

We recruit 28 user interface (UI) and user experience (UX) designers with experience in layout design for the user study sessions. Each participant is presented with 48 randomly generated questions. In each question, there will be a pair of layouts, and the user needs to pick the better one (Figure \ref{user_test_q}. Each layout pair is selected from two of these three groups: ground truth, VTN, and PLay. We also ensure the 48 questions equally cover all possible combinations of these groups. 

When answering a question, the group gets +1 score if the user pick the layout that belongs to this group, and gets -1 vice versa. Both groups get 0 if the user thinks their layouts are equally good or bad. The final scores are normalized by the number of questions.

We also give the following criteria and information to the participants:

\begin{itemize}
\item Please evaluate for both aesthetics and functionalities of the layouts. For example, the alignments, proportions, how reasonable for the buttons to be put here, etc.

\item Some of the layouts are intentionally not valid nor optimal (many of them are synthesized). Therefore, please do not try too hard to justify every layout. Use your intuition and experience as a designer to pick the better one from each pair.

\item Some of the examples that do not look like a full mobile UI screen might still be valid designs. They can represent UI cases such as popped windows, opened drawers, or simply without background image.

\item Note that the text elements are often not aligned due to their various lengths.
\end{itemize}

\section{First-stage Models}\label{fs_appendix}
In Table \ref{fs}, we can see that $\beta$ values do not have significant effect on the metrics as long as they are small enough ($<=0.1$). Latent dimension $d=8$ seems to be the optimal choice in our experiments. VQVAEs achieve comparable numbers in FID and FD-VG but fall short on G-Usage. Although increasing the codebook size improves G-Usage, it is less computationally efficient to use $c>16384$ compared to a simple VAE.

\begin{table}[t]
\caption{List of first-stage models. $\beta$ is the weight of KL loss, $d$ is the dimension of the latent space for each element, and $c$ is the codebook size of VQVAE. In each row, FID-Recon and FD-VG-Recon are computed using the reconstruction samples generated by the first-stage model. FID, FD-VG, and G-Usage are computed using PLay trained on this first-stage model.}
\label{fs}
\vskip 0.15in
\begin{center}
\begin{small}
\begin{tabular}{l|ccc|cc|ccc}
\toprule
First-stage & $\beta$ & $d$ & $c$ & FID-Recon & FD-VG-Recon & FID & FD-VG & G-Usage\\
\midrule
None & n/a & 8 & n/a & n/a & n/a & 167.9 & 4.683 & \textbf{0.991*}\\
\midrule
VAE & 1.0 & 8 & n/a & 19.75 & 2.04$e^{-1}$ & 14.35 & 0.311 & 0.835\\
& 1$e^{-1}$ & 8 & n/a & 4.010 & 1.94$e^{-2}$ & 11.73 & 0.279 & 0.955\\
& 1$e^{-2}$ & 8 & n/a & 0.612 & 3.99$e^{-3}$ & 11.63 & 0.262 & 0.968\\
& 1$e^{-3}$ & 8 & n/a & 0.215 & 1.75$e^{-3}$ & \textbf{10.59} & 0.245 & 0.964\\
& 1$e^{-4}$ & 8 & n/a & 0.142 & 1.35$e^{-3}$ & 11.49 & 0.252 & 0.964\\
& 5$e^{-4}$ & 8 & n/a & 0.138 & 1.49$e^{-3}$ & 11.57 & \textbf{0.236} & 0.965\\
& 1$e^{-5}$ & 4 & n/a & 3.805 & 2.02$e^{-2}$ & 12.76 & 0.281 & 0.946\\
& 1$e^{-5}$ & 8 & n/a & 0.115 & 1.46$e^{-3}$ & 11.05 & 0.253 & 0.965\\
& 1$e^{-5}$ & 16 & n/a & 0.053 & 1.30$e^{-3}$ & 11.58 & 0.257 & 0.953\\
\midrule
VQVAE & n/a & 8 & 1024 & 10.70 & 6.52$e^{-2}$ & 11.34 & 0.247 & 0.909\\
& n/a & 8 & 4096 & 8.68 & 5.22$e^{-2}$ & 11.45 & 0.244 & 0.924\\
& n/a & 8 & 16384 & 6.68 & 3.70$e^{-2}$ & 11.49 & 0.254 & 0.937\\
\bottomrule
\end{tabular}
\end{small}
\end{center}
\footnotesize{$^*$PLay without the first-stage model generates mostly random, unaligned boxes and therefore, has a high guideline usage.}
\vskip -0.1in
\end{table}

\section{Layout Inpainting Details}
We generate layout inpainting results following the steps below:
\begin{enumerate}
    \item Given a layout with $k$ elements, mask out $n$ elements within an area with indices $idx_{mask} = [m_1, m_2, ..., m_n]$.
    \item Encode the layout in Step $1$.
    \item Apply forward diffusion process for the encoding from Step $2$ to get its latent embeddings at each time step $t$.
    \item Start the diffusion sampling process with $k$ elements. At time step $T$, use the corresponding embedding generated in Step $3$, and swap the embeddings at $idx_{mask}$ with noise $z$. Then compute the embeddings at $T-1$ using the backward diffusion process.
    \item At each time step $t<T$, swap the generated embeddings from time step $t+1$ with the corresponding embeddings from Step $3$ at all indices except from $idx_{mask}$. Then compute the embeddings at $t-1$ and repeat this step until $t=0$.
    \item Decode the final embeddings back to a layout, extract its elements at $idx_{mask}$, and swap the corresponding elements in the original layout with the extracted ones.
\end{enumerate}

\section{Color Legend}
The color legend for CLAY can be found in Table \ref{clay_legend}, and the color legend for RICO-Semantic and PublayNet can be found in Tabel \ref{rico_publaynet_legend}. The border (stroke) color is \#393e46 with stroke width $1$ for all boxes across the datasets and generated results.

\begin{table}[t]
\caption{Color legend for rendering the CLAY dataset and the generated results.}
\label{clay_legend}
\vskip 0.15in
\begin{center}
\begin{small}
\begin{tabular}{lc|c}
\toprule
Class & Index & Color\\
\midrule
IMAGE & $0$ & \#a6e3e9\\
PICTOGRAM & $1$ & \#bad7df\\
BUTTON & $2$ & \#71c9ce\\
TEXT & $3$ & \#cbf1f5\\
LABEL & $4$ & \#dbe2ef\\
TEXT\_INPUT & $5$ & \#f6f6f6\\
MAP & $6$ & \#e3fdfd\\
CHECK\_BOX & $7$ & \#ffe2e2\\
SWITCH & $8$ & \#ffd3b6\\
PAGER\_INDICATOR & $9$ & \#b4846c\\
SLIDER & $10$ & \#8785a3\\
RADIO\_BUTTON & $11$ & \#c06c84\\
SPINNER & $12$ & \#f38181\\
PROGRESS\_BAR & $13$ & \#dcd6f7\\
ADVERTISEMENT & $14$ & \#364f6b\\
DRAWER & $15$ & \#d3e0dc\\
NAVIGATION\_BAR & $16$ & \#3f72af\\
TOOLBAR & $17$ & \#a6b1e1\\
LIST\_ITEM & $18$ & \#bbded6\\
CARD\_VIEW & $19$ & \#ffb6b9\\
CONTAINER & $20$ & \#fae3d9\\
DATE\_PICKER & $21$ & \#99ddcc\\
NUMBER\_STEPPER & $22$ & \#7d5a50\\
\bottomrule
\end{tabular}
\end{small}
\end{center}
\vskip -0.1in
\end{table}

\begin{table}[t]
\caption{Color legend for rendering the RICO-Semantic dataset, PublayNet dataset, and the generated results.}
\label{rico_publaynet_legend}
\vskip 0.15in
\begin{center}
\begin{small}
\begin{tabular}{lc|lc|c}
\toprule
RICO-Semantic & & PublayNet & & \\
\midrule
Class & Index & Class & Index & Color\\
\midrule
TEXT & $0$ & TEXT & $0$ & \#cbf1f5\\
LIST\_ITEM & $1$ & TITLE & $1$ & \#bbded6\\
IMAGE & $2$ & LIST & $2$ & \#a6e3e9\\
TEXT\_BUTTON & $3$ & TABLE & $3$ & \#71c9ce\\
ICON & $4$ & FIGURE & $4$ & \#fae3d9\\
TOOLBAR & $5$ & & &\#a6b1e1\\
TEXT\_INPUT & $6$ & & &\#f6f6f6\\
ADVERTISEMENT & $7$ & & &\#364f6b\\
CARD\_VIEW & $8$ & & &\#ffb6b9\\
WEB\_VIEW & $9$ & & &\#f38181\\
DRAWER & $10$ & & &\#d3e0dc\\
BACKGROUND\_IMAGE & $11$ & & &\#e3fdfd\\
RADIO\_BUTTON & $12$ & & &\#c06c84\\
MODAL & $13$ & & &\#dcd6f7\\
MULTI\_TAB & $14$ & & &\#ea5455\\
PAGER\_INDICATOR & $15$ & & &\#dbe2ef\\
SLIDER & $16$ & & &\#3f72af\\
SWITCH & $17$ & & &\#bad7df\\
MAP & $18$ & & &\#ffd3b6\\
BOTTO\_NAVIGATION & $19$ & & &\#b4846c\\
VIDEO & $20$ & & &\#8785a3\\
CHECK\_BOX & $21$ & & &\#99ddcc\\
BUTTON\_BAR & $22$ & & &\#7d5a50\\
NUMBER\_STEPPER & $23$ & & &\#ffd460\\
DATE\_PICKER & $24$ & & &\#f07b3f\\
\bottomrule
\end{tabular}
\end{small}
\end{center}
\vskip -0.1in
\end{table}

\section{More Results}
\begin{figure}[ht]
\begin{center}
\centerline{\includegraphics[width=0.9\columnwidth]{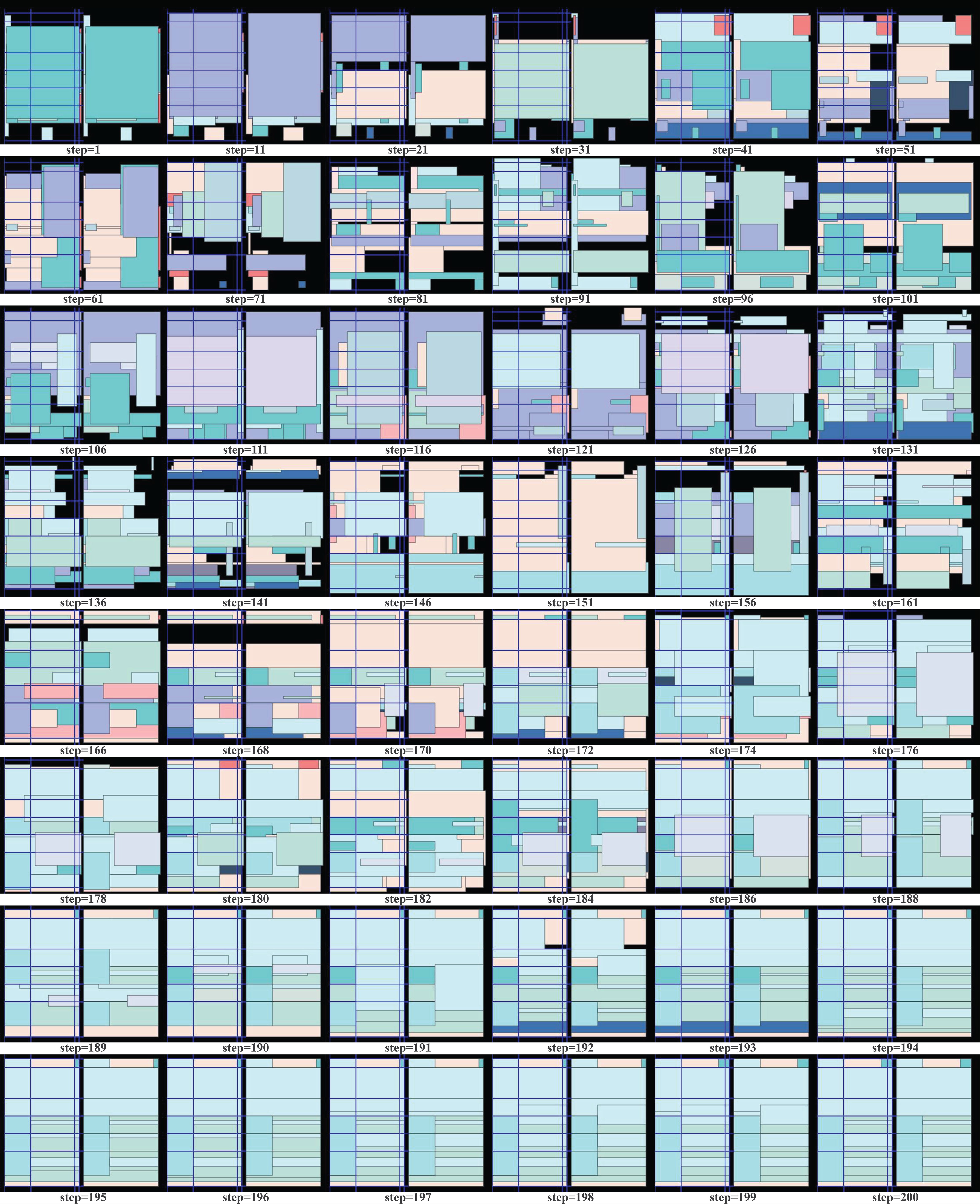}}
\caption{Decoding the latent diffusion steps back to layouts. Interestingly, although the first 150 steps look noisy and random, the last 50 steps still demonstrate the nature of the denoising process, forming from low frequency features to high frequency features. This coarse-to-fine generation process can be commonly found in imaged-based diffusion models, but it is the first time we observe the same process in latent diffusion models for vector graphics.}
\label{steps}
\end{center}
\vskip -0.2in
\end{figure}

\begin{figure}[ht]
\begin{center}
\centerline{\includegraphics[width=0.4\columnwidth]{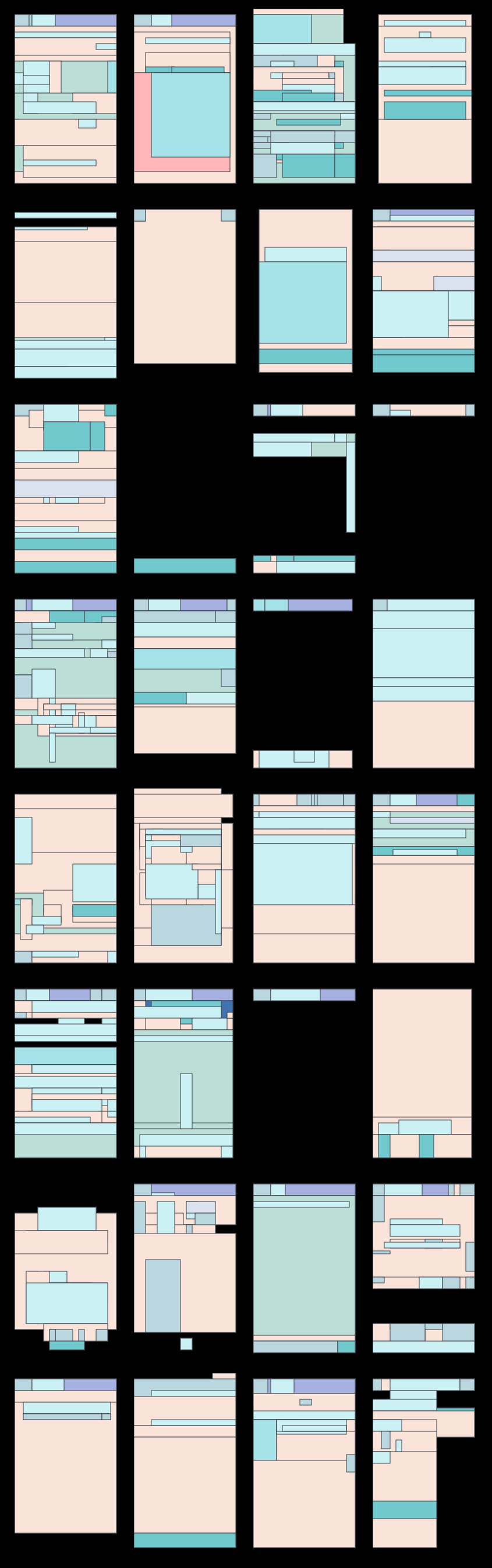}}
\caption{Samples generated by VTN trained on CLAY.}
\label{samples4}
\end{center}
\vskip -0.2in
\end{figure}

In Figure \ref{steps}, we decode the output from each denoising step in the sampling process and visualize the rendered results from them. We also visualize the generated samples from PLay (Figure \ref{samples1}, \ref{samples2}, and \ref{samples3}) and from VTN (Figure \ref{samples4}).

\begin{figure}[ht]
\begin{center}
\centerline{\includegraphics[width=0.7\columnwidth]{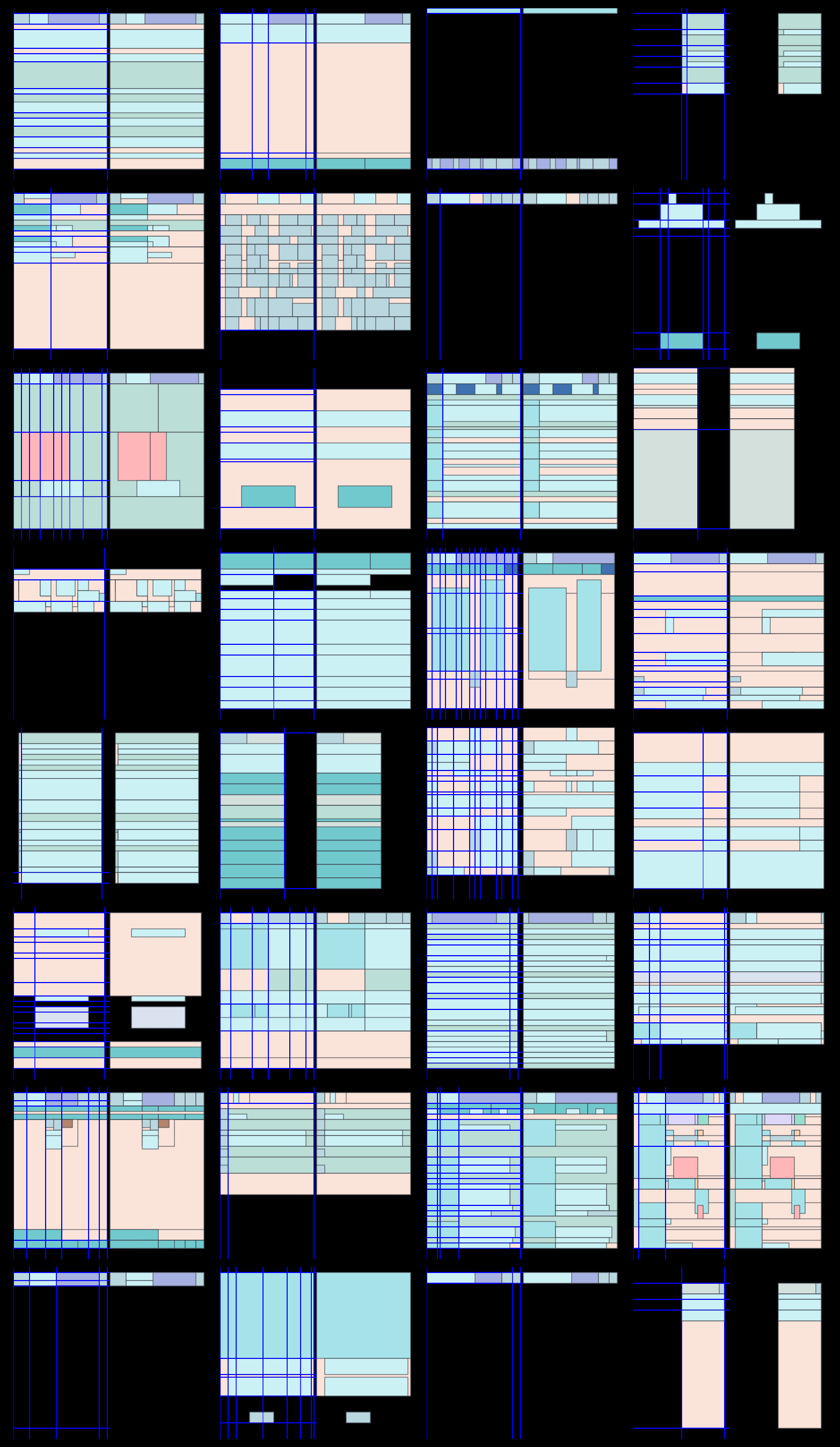}}
\caption{Samples generated by PLay trained on CLAY. In each cell, left: input guidelines (blue lines) on top of the generated layout. Right: generated layout.}
\label{samples1}
\end{center}
\vskip -0.2in
\end{figure}

\begin{figure}[ht]
\begin{center}
\centerline{\includegraphics[width=0.7\columnwidth]{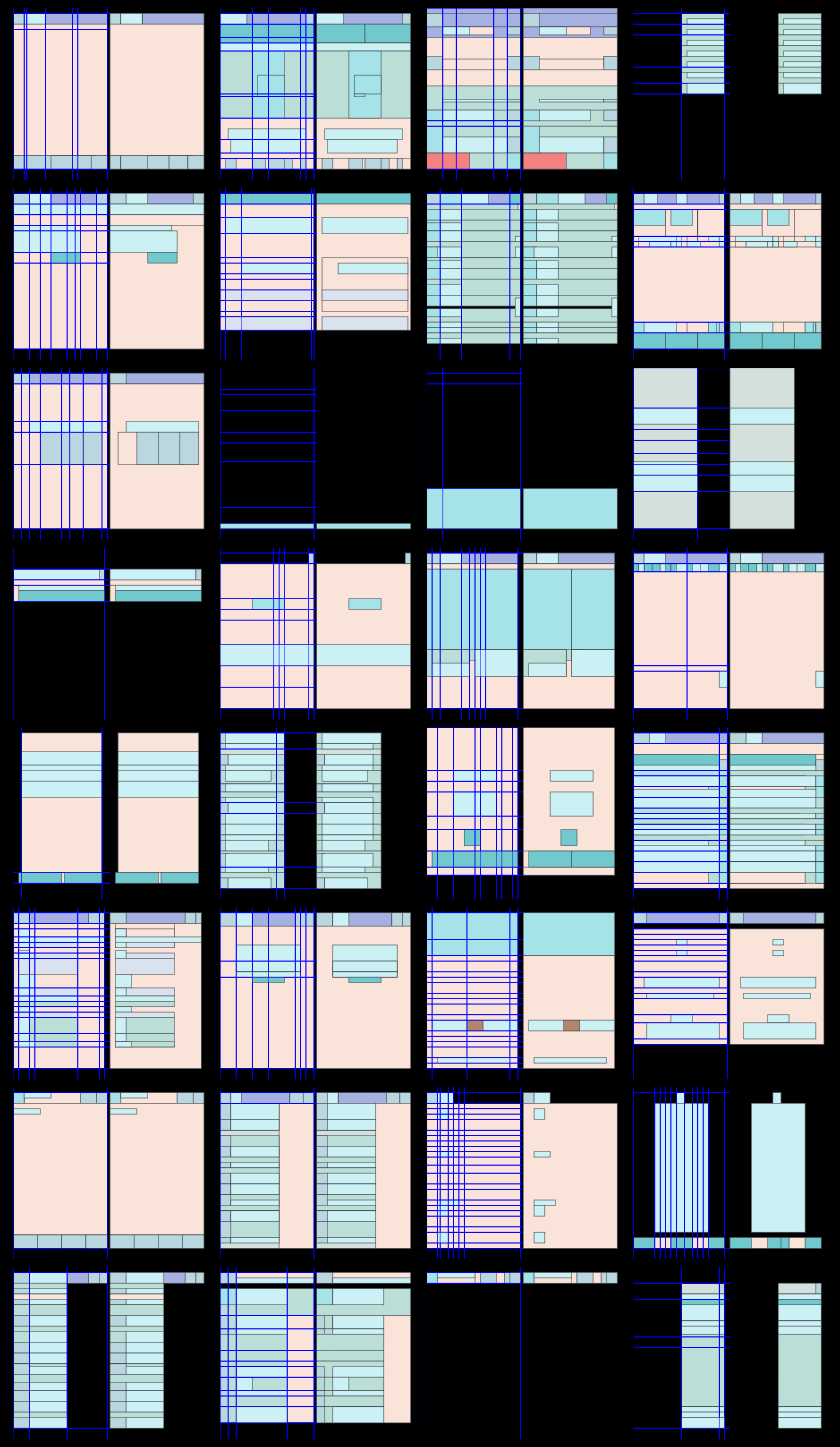}}
\caption{(continued) Samples generated by PLay trained on CLAY.}
\label{samples2}
\end{center}
\vskip -0.2in
\end{figure}

\begin{figure}[ht]
\begin{center}
\centerline{\includegraphics[width=0.7\columnwidth]{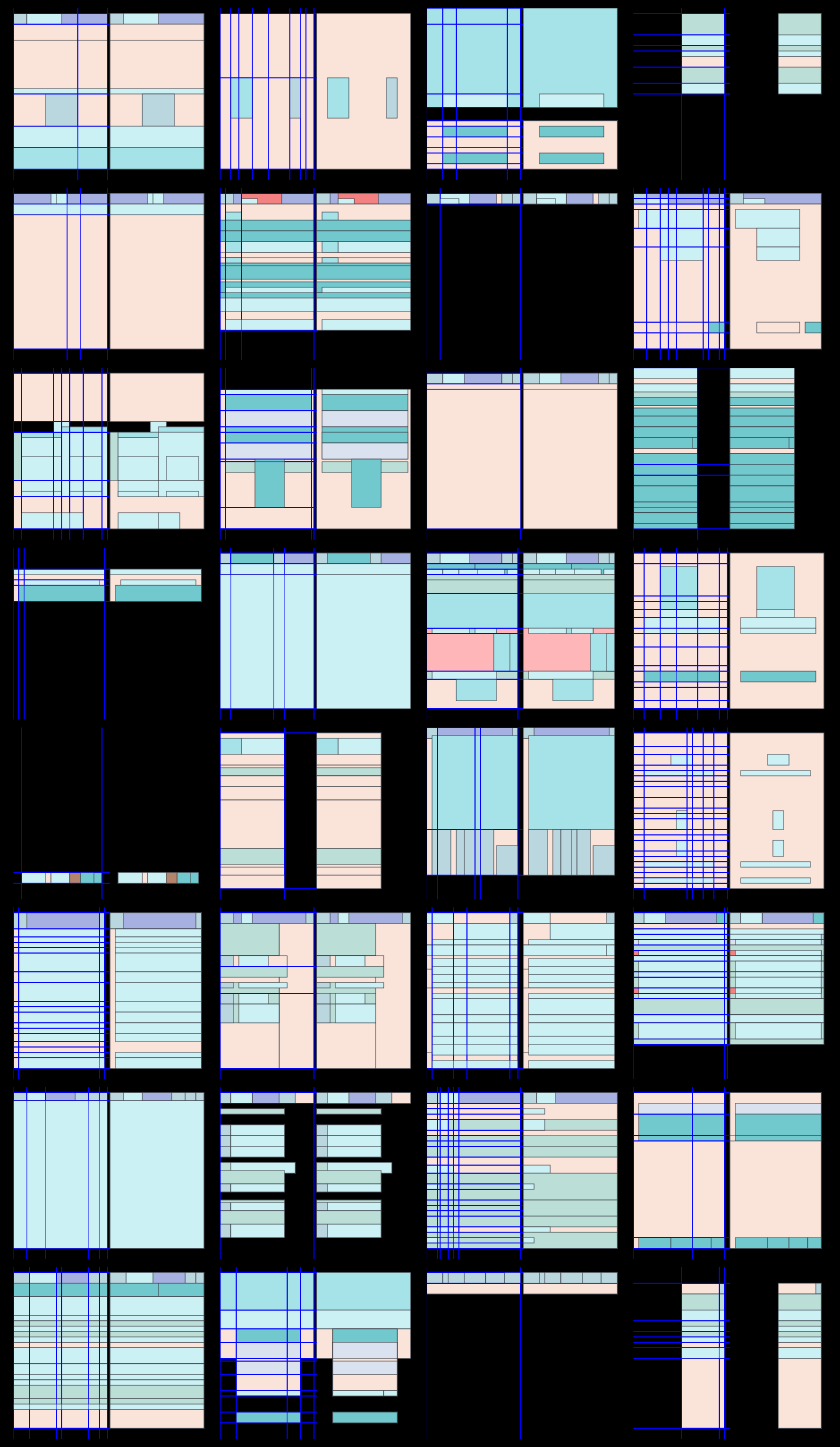}}
\caption{(continued) Samples generated by PLay trained on CLAY.}
\label{samples3}
\end{center}
\vskip -0.2in
\end{figure}

\end{document}